\documentclass[10pt,twocolumn,letterpaper]{article}

\usepackage{cvpr}      

\usepackage{graphicx}
\usepackage{amsmath}
\usepackage{amssymb}
\usepackage{booktabs}

\usepackage[pagebackref,breaklinks,colorlinks]{hyperref}

\usepackage[capitalize]{cleveref}
\crefname{section}{Sec.}{Secs.}
\Crefname{section}{Section}{Sections}
\Crefname{table}{Table}{Tables}
\crefname{table}{Tab.}{Tabs.}

\usepackage{bbm}
\usepackage{bbding}
\usepackage{caption}    
\usepackage{subcaption} 
\usepackage{pifont}
\usepackage{enumitem}
\usepackage{wasysym}
\usepackage{algorithmic}
\usepackage{algorithm}
\usepackage{multirow}
\usepackage[table]{xcolor}
\definecolor{mygray}{gray}{0.9}

\definecolor{iconblue}{rgb}{0, 0.60, 0.80}

\def\cvprPaperID{8834} 
\def\confName{CVPR}
\def\confYear{2023}

\begin{document}

\title{Unbiased Multiple Instance Learning for Weakly Supervised \\Video Anomaly Detection}

\author{Hui Lv$^{1,3,4}$, Zhongqi Yue$^4$, Qianru Sun$^3$, Bin Luo$^2$, Zhen Cui$^1$\thanks{Corresponding author}, Hanwang Zhang$^4$\\
$^1$PCAlab, Nanjing University of Science and Technology\quad $^2$Alibaba Group\\
$^3$Singapore Management University\quad $^4$Nanyang Technological University\\
{\tt\small $^1$\{hubrthui, zhen.cui\}@njust.edu.cn, $^2$luwu.lb@alibaba-inc.com}\\
{\tt\small$^3$qianrusun@smu.edu.sg, $^4$\{yuez0003, hanwangzhang\}@ntu.edu.sg}
}

\maketitle
\begin{abstract}
   Weakly Supervised Video Anomaly Detection (WSVAD) is challenging because the binary anomaly label is only given on the video level, but the output requires snippet-level predictions. So, Multiple Instance Learning (MIL) is prevailing in WSVAD. However, MIL is notoriously known to suffer from many false alarms because the snippet-level detector is easily biased towards the abnormal snippets with simple context, confused by the normality with the same bias, and missing the anomaly with a different pattern. To this end, we propose a new MIL framework: Unbiased MIL (UMIL), to learn unbiased anomaly features that improve WSVAD. At each MIL training iteration, we use the current detector to divide the samples into two groups with different context biases: the most confident abnormal/normal snippets and the rest ambiguous ones. Then, by seeking the invariant features across the two sample groups, we can remove the variant context biases. 
   Extensive experiments on benchmarks UCF-Crime and TAD demonstrate the effectiveness of our UMIL. Our code is provided at this \href{https://github.com/ktr-hubrt/UMIL}{link}.
\end{abstract}
\section{Introduction}
\label{sec:intro}
Video Anomaly Detection (VAD) aims to detect events among video sequences that deviate from expectation, which is widely applied in real-world tasks such as intelligent manufacturing~\cite{Huang2022Survey}, TAD surveillance~\cite{kamijo2000traffic,lv2021localizing} and public security~\cite{mohammadi2016angry,sultani2018real}. To learn such a detector, conventional fully-supervised VAD~\cite{acsintoae2022ubnormal} is impractical as the scattered but diverse anomalies require extremely expensive labeling cost. On the other hand, unsupervised VAD~\cite{zhao2011online,kratz2009anomaly,wu2010chaotic,li2013anomaly,antic2011video} by only learning on normal videos to detect open-set anomalies often triggers false alarms, as it is essentially ill-posed to define what is normal and abnormal by giving only normal videos without any prior knowledge. Hence, we are interested in a more practical setting: Weakly Supervised VAD (WSVAD)~\cite{zhong2019graph,li2022self}, where only video-level binary labels (\ie, normal \textit{vs.} abnormal) are available.

\begin{figure}[t]
    \centering
    \footnotesize
    \begin{subfigure}[t]{0.23\textwidth}
         \includegraphics[width=\textwidth]{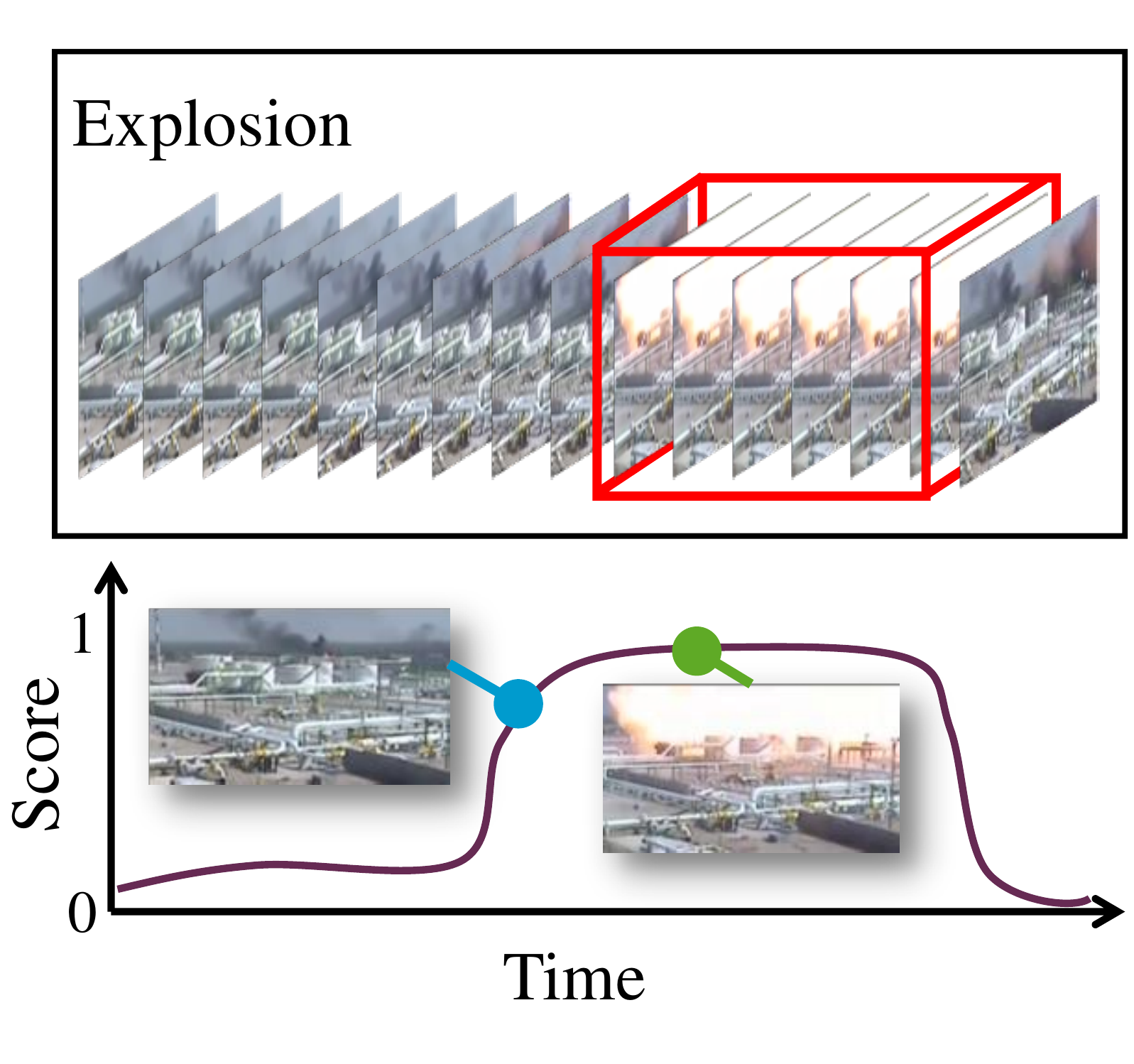}
         \caption{}
         \label{fig:1a}
    \end{subfigure}
    \begin{subfigure}[t]{0.23\textwidth} 
         \includegraphics[width=\textwidth]{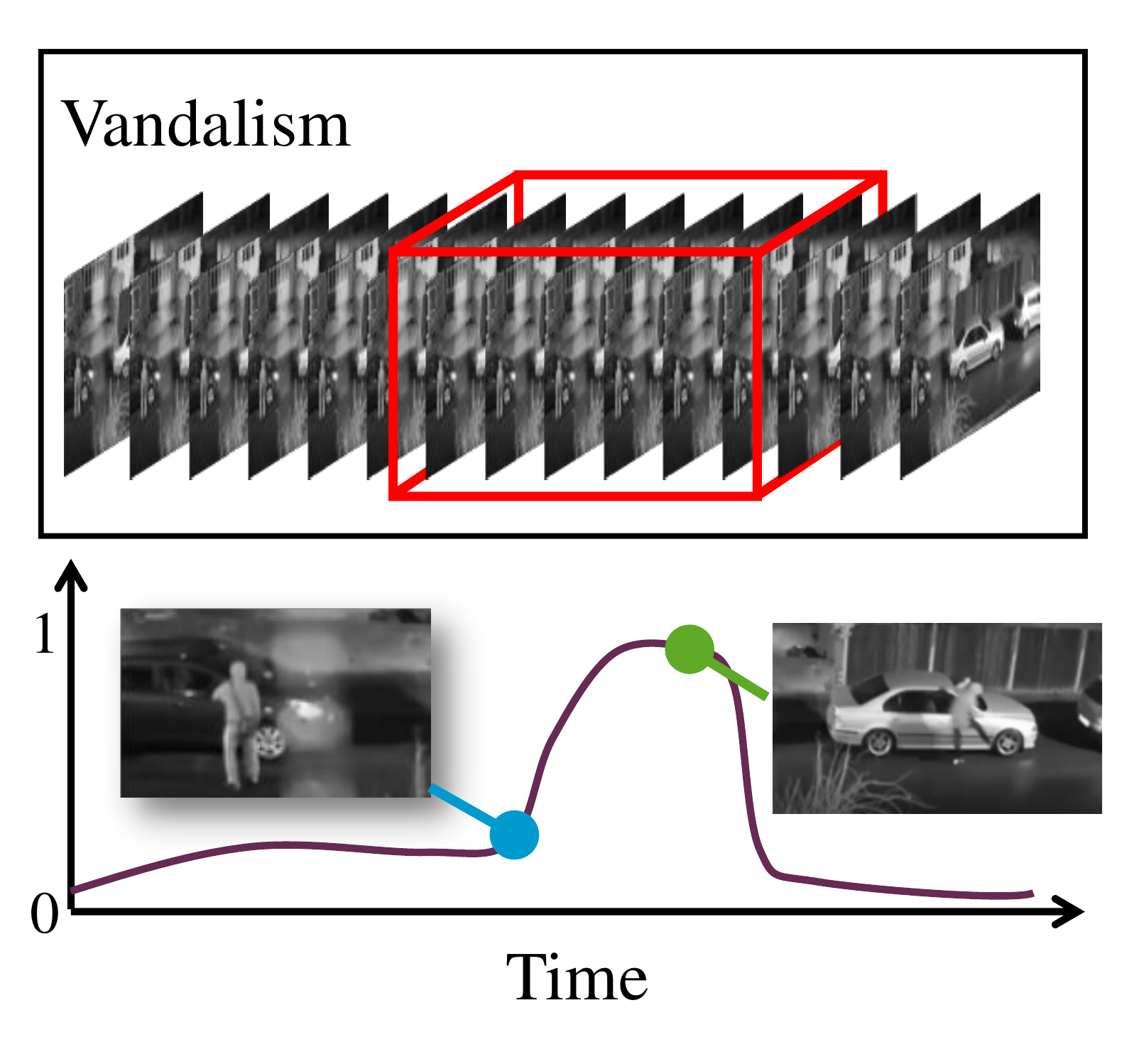}
         \caption{}
         \label{fig:1b}   
    \end{subfigure}
    \vspace{-2mm}
	\caption{Two anomalies of Explosion and Vandalism are illustrated. Among each video sequence, we use red boxes to highlight the ground-truth anomaly regions as in the first row. The corresponding anomaly curves of an MIL-based model are depicted below. False alarms and real anomalies are linked to the curves with blue arrows and green arrows respectively. Best viewed in color.}
    \vspace{-6mm}
	\label{fig:abstract}
\end{figure}

In WSVAD, each video sequence is partitioned into multiple snippets. Hence, all the snippets are normal in a normal video, and at least one snippet contains the anomaly in an abnormal one. The goal of WSVAD is to train a snippet-level anomaly detector using video-level labels. The mainstream method is Multiple Instance Learning (MIL)~\cite{lv2021localizing,sultani2018real}---multiple instances refer to the snippets in each video, and learning is conducted by decreasing the predicted anomaly score for each snippet in a normal video, and increasing that only for  the snippet with the largest anomaly score in an abnormal video. For example, Figure~\ref{fig:1a} shows an abnormal video containing an explosion scene, and the detector is trained by MIL to increase the anomaly score for the most anomalous explosion snippet (green link).

However, MIL is easily biased towards the simplest context shortcut in a video. We observe in Figure~\ref{fig:1a} that the detector is biased to smoke, as the pre-explosion snippet with only smoke is also assigned a large anomaly score (blue link). This biased detector can trigger false alarms on smoke snippets without anomaly, \eg, a smoking chimney. Moreover, it could also fail in videos with multiple anomalies of different contexts. In Figure~\ref{fig:1b}, the video records two men vandalizing a car, where only the second one has substantial motions. We notice that the two snippets of them have large differences in the anomaly scores, and only the latter is predicted as an anomaly. This shows that the detector is biased to the drastic motion context while being less sensitive to the subtle vandalism behavior, which is the true anomaly.

\begin{figure}
    \centering
    \footnotesize
    \begin{subfigure}[t]{\linewidth}
         \includegraphics[width=\textwidth]{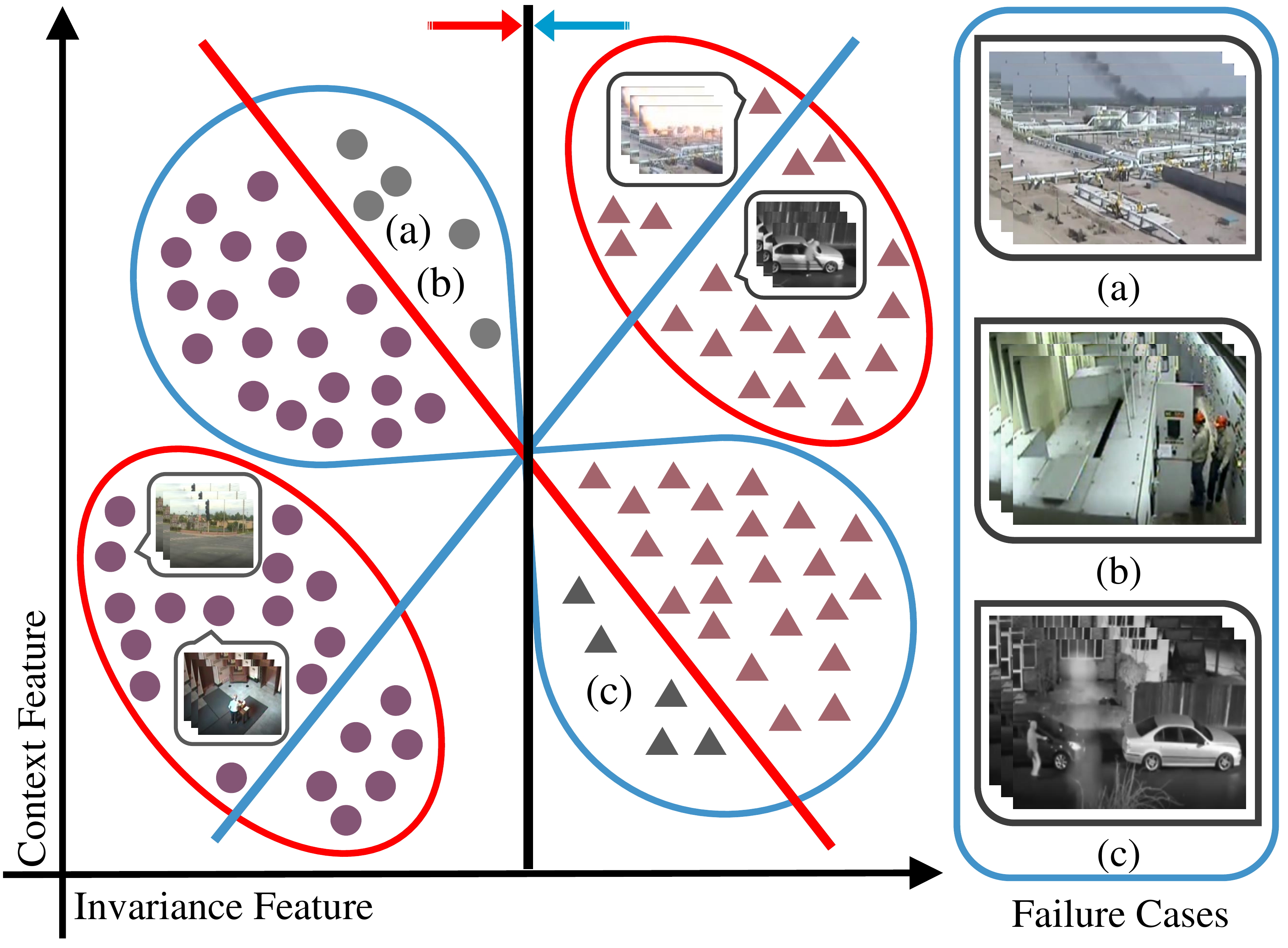}
         \phantomcaption
         \label{fig:2a}
    \end{subfigure}
    \begin{subfigure}[t]{0\textwidth} 
         \includegraphics[width=\textwidth]{example-image-b}
         \phantomcaption
         \label{fig:2b}   
    \end{subfigure}
    \begin{subfigure}[t]{0\textwidth} 
         \includegraphics[width=\textwidth]{example-image-b}
         \phantomcaption
         \label{fig:2c}   
    \end{subfigure}
    \vspace*{-8mm}
    \caption{\textcolor{red}{Red}: Confident Set, \textcolor{iconblue}{Blue}: Ambiguous Set. {\scriptsize \CIRCLE}: Normal sample, $\blacktriangle$: Abnormal sample, \textcolor{gray}{Gray instances}: Failure cases. The red line denotes the classifier trained under MIL. The invariant classifier (black line) can be learned by combining confident snippets learning in MIL (red line) and the ambiguous snippets clustering (blue line). Best viewed in color.}
    \vspace{-6mm}
    \label{fig:2}
\end{figure}

The root of MIL's biased predictions lies in its training scheme with biased sample selection. As shown in Figure~\ref{fig:2}, the bottom-left cluster (denoted as the red ellipse) corresponds to the confident normal snippets, \eg, an empty crossroad or an old man standing in a room, which are either from normal videos as the ground truth or from abnormal videos but visually similar to the ground-truth ones. On the contrary, the top-right cluster denotes the confident abnormal ones, which not only contain the true anomaly features (\eg, explosion and vandalism) but also include the context features commonly appearing with anomaly under a context bias (\eg, smoke and motions). In MIL, the trained detector is dominated by the confident samples, corresponding to the top-right cluster with the abnormal representation and the bottom-left cluster with the normal representation. Hence the learned detector (red line) inevitably captures the context bias in the confident samples. Consequently, the biased detector generates ambiguous predictions on snippets with a different context bias (the red line mistakenly crossing the blue points), \eg, smoke but normal (industrial exhaust in Figure~\ref{fig:2a}), substantial motion but normal (equipment maintenance in Figure~\ref{fig:2b}), or subtle motion but abnormal (vandalizing the rear-view mirror in Figure~\ref{fig:2c}), leading to the aforementioned failure cases.

To this end, we aim to build an unbiased MIL detector by training with both the confident abnormal/normal and the ambiguous ones.
Specifically, at each UMIL training iteration, we divide the snippets into two sets using the current detector: 1) the confident set with abnormal and normal snippets and 2) the ambiguous set with the rest snippets, \eg, the two sets are enclosed with red circles and blue circles in Figure~\ref{fig:2}, respectively.
The ambiguous set is grouped into two unsupervised clusters (\eg, the two blue circles separated by the blue line) to discover the intrinsic difference between normal and abnormal snippets.
Then, we seek an invariant binary classifier between the two sets that separate the abnormal/normal in the confident set and the two clusters in the ambiguous one. The rationale of the proposed invariance pursuit is that the snippets in the ambiguous set must have a different context bias from the confident set, otherwise, they will be selected into the same set.
Therefore, given a different context but the same true anomaly, the invariant pursuit will turn to the true anomaly (\eg, the black line).

Overall, we term our approach as \textbf{Unbiased MIL (UMIL)}.
Our contributions are summarized below:
\begin{itemize}[leftmargin=+0.1in,itemsep=5pt,topsep=5pt,parsep=0pt]
    \item UMIL is a novel WSVAD method that learns an unbiased anomaly detector by pursuing the invariance across the confident and ambiguous snippets with different context biases.
    \item Thanks to the unbiased objective, UMIL is the first WSVAD method that combines feature fine-tuning and detector learning into an end-to-end training scheme. This leads to a more tailored feature representation for VAD.
    \item UMIL is equipped with a fine-grained video partitioning strategy for preserving the subtle anomaly information in video snippets.
    \item These contribute to the improved performance over the current state-of-the-art methods on UCF-Crime~\cite{sultani2018real} ( $1.4\%$ AUC) and TAD~\cite{lv2021localizing} ($3.3\%$ AUC) benchmarks. Note that UMIL brings more than 2\% AUC gain compared with the MIL baseline on both datasets, which justifies the effectiveness of UMIL.
\end{itemize}
\section{Related Work}
\label{sec:RW}
The research lineup of video anomaly detection falls into two classes:  unsupervised and weakly-supervised settings.

\noindent\textbf{Unsupervised methods} include the ones that only use unlabelled training data or directly train and test on testing data. Del~\etal~\cite{del2016discriminative} proposed to detect changes on a sequence of video data to detect unique frames. Tudor~\etal~\cite{tudor2017unmasking} introduced unmasking technology~\cite{koppel2007measuring} to iteratively train a binary classifier to distinguish the most discriminant features. Lately, Zaheer~\etal~\cite{zaheer2022generative} exploited the low frequency of anomalies by building a cross-supervision between a generator and a discriminator.
There are also One-Class Classification (OCC) methods assume the availability of normal training data only and approach the problem in an unsupervised manner. 
Typically, researchers fit a model with only normal data, then detect anomalies by distinguishing the events that deviate from the model. 
Early works used hand-crafted appearance and motion features~\cite{adam2008robust,antic2011video,lu2013abnormal,mahadevan2010anomaly,mehran2009abnormal}.
Thanks to the impressive progress of deep learning, recent works used the features from pre-trained deep neural networks and built an anomaly classifier upon them~\cite{ravanbakhsh2018plug,hasan2016learning}. There are also methods for self-supervised feature learning~\cite{sabokrou2015real,xu2015learning}, where a popular approach is by temporal prediction~\cite{xingjian2015convolutional,liu2018future,lv2021learning}.
However, unsupervised methods suffer from false alarms for unseen normal patterns, since it is impossible to collect all kinds of normality in one dataset.

\noindent\textbf{Weakly-supervised methods} exploit both normal and abnormal training data with weak annotations only on the video-level~\cite{sultani2018real}. 
Multiple instance learning (MIL) is the mainstream paradigm that uses video-level labels for training snippet-level anomaly detectors~\cite{sultani2018real,he2018anomaly,zhu2019motion}. 
Generally, they embrace the two-stage anomaly detection pipeline, which performs anomaly detection upon pre-extracted features.
In particular, Zhong~\etal~\cite{zhong2019graph} considered the WSVAD task as supervised learning under noise labels and they designed an alternate training procedure to enhance the discrimination of action classifiers.
Lv~\etal~\cite{lv2021localizing} focused on anomaly localization and proposed a higher-order context model as well as a margin-based MIL loss.
Tian~\etal~\cite{tian2021weakly} investigated the feature magnitude to facilitate anomaly detection and selected the instances of top-k scores to better represent the video for MIL.
Li~\etal~\cite{li2022self} proposed multiple sequence learning, where consecutive snippets with high anomaly scores are selected in MIL learning.
They attempted to improve the sample selection for improving MIL, whose biased nature is not changed yet. In this paper, our unbiased MIL framework is the first effort on removing the context bias~\cite{yue2020interventional,yue2021transporting} in WSVAD. In addition, we integrate feature representation fine-tuning and anomaly detector learning into an end-to-end training fashion.
\section{Method}
\label{sec:Med}

\begin{figure*}[t]
	\centering
	\includegraphics[width=1\textwidth]{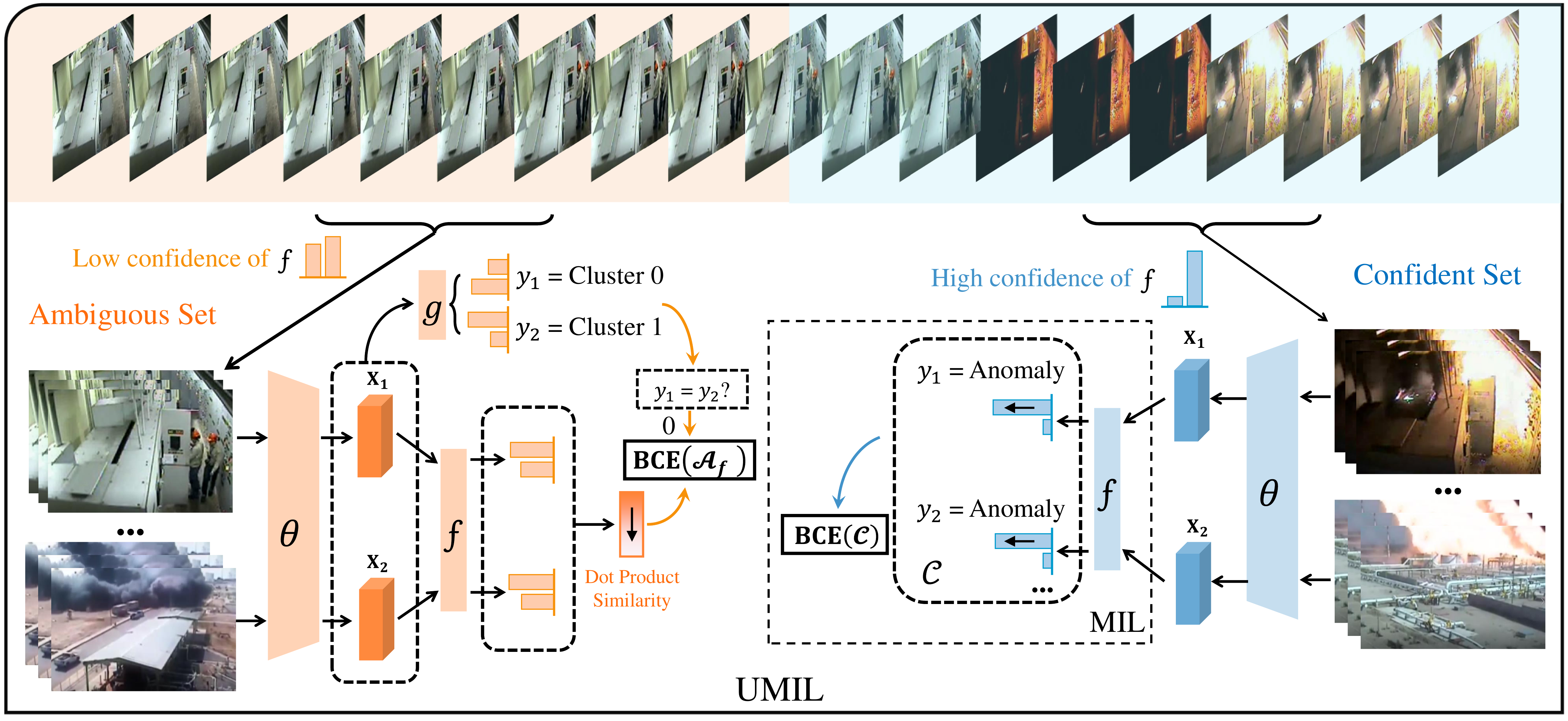}
    \vspace{-6mm}
	\caption{The proposed UMIL framework for WSVAD consists of a backbone model $\theta$, an anomaly head $f$, and a cluster head $g$. We use the predictions by $f$ to divide the snippets into a confident set $\mathcal{C}$ and an ambiguous set $\mathcal{A}$. In MIL, the model is only supervised by the confident snippets to further increase the confidence of anomaly prediction (the black arrows on the probability bar). In UMIL, $f$ is additionally supervised by $\mathcal{A}$ to separate the two clusters identified by $g$ for removing the context bias in $\mathcal{C}$. The black arrow on the similarity bar denotes that minimizing the BCE losses in $\mathcal{A}$ will decrease the dot-product similarity of the predictions on the pair, as they are from different clusters ($y_1 \neq y_2$). Best viewed in color.}
	\label{fig:framework}
    \vspace{-4mm}
\end{figure*}

In Weakly Supervised Video Anomaly Detection (WSVAD), each training video is annotated with a binary anomaly label $y \in \{0, 1\}$ (\ie, normal or abnormal) and partitioned into $m$ snippets. We denote $\mathbf{x}_i, i\in\{1,\ldots, m\}$ as the feature of the $i$-th snippet in the video extracted by a backbone parameterized by $\theta$. The goal of WSVAD is to train a snippet-level anomaly classifier $f(\mathbf{x}_i)$ predicting the probability of the snippet being positive (abnormal).

\subsection{From MIL to Unbiased MIL}

The mainstream method in WSVAD is Multiple Instance Learning (MIL). In MIL, the backbone $\theta$ is pre-trained (\eg, on Kinetics400~\cite{carreira2017quo}) and frozen in training. It aims to learn $f$ so as to predict the most anomalous snippet in a normal video (\ie, $y=0$) as normal, and that in an abnormal video (\ie, $y=1$) as abnormal.
Specifically, for each video, MIL creates a tuple containing the prediction of $f$ on the most anomalous snippet and the video's anomaly label, \ie, $(\mathrm{max}\{f(\mathbf{x}_i)\}_{i=1}^m, y)$. Then MIL aggregates the tuple for all videos to construct a labeled confident snippet set $\mathcal{C}$, and trains $f$ by minimizing the binary cross-entropy (BCE) loss:
\begin{equation}
    \textrm{BCE}(\mathcal{C}) = -\mathop{\mathbb{E}}_{(\hat{y},y)\sim \mathcal{C}} \left[ y\mathrm{log} (\hat{y}) + (1-y)\mathrm{log}(1-\hat{y}) \right],
    \label{eq:1}
\end{equation}
where $\hat{y} = \mathrm{max}\{f(\mathbf{x}_i)\}_{i=1}^m$. Note that some methods~\cite{sultani2018real} use the mean squared error loss, which achieves the same outcome as Eq.~\eqref{eq:1}.
In this way, for a normal video with $y=0$, by \emph{minimizing} $\mathrm{max}\{f(\mathbf{x}_i)\}_{i=1}^m$, $f$ must assign low abnormal probability for all the snippets. For an abnormal video with $y=1$, by \emph{maximizing} $\mathrm{max}\{f(\mathbf{x}_i)\}_{i=1}^m$, $f$ is trained to output an even larger probability for the most confident abnormal snippet.
However, the MIL training scheme suffers from biased sample selection: as $f$ is trained to further increase $\mathrm{max}\{f(\mathbf{x}_i)\}_{i=1}^m$ in an abnormal video, the rest ambiguous snippets become even less likely to be selected by $\mathrm{max}$. Hence MIL essentially discards the ambiguous snippets and only trains on the confident ones, which leads to a biased detector (\eg, Figure~\ref{fig:2}).

In contrast, our proposed Unbiased MIL (UMIL) leverages both the confident and ambiguous snippets to train the anomaly classifier $f$. Specifically, in Step 1, we divide the snippets into a labeled confident snippet set $\mathcal{C}$ and an unlabeled ambiguous snippet set $\mathcal{A}$. In Step 2, we cluster $\mathcal{A}$ into 2 groups in an unsupervised fashion to distinguish the normal and abnormal snippets. Finally, in Step 3, $f$ is supervised by both $\mathcal{C}$ and $\mathcal{A}$ to simultaneously predict the binary labels in $\mathcal{C}$ and separate the clusters in $\mathcal{A}$.

\subsection{Step 1: Divide Snippets}
\label{sec:step1}
Based on the predictions from $f$, we divide the snippets into the confident set $\mathcal{C}$ and the ambiguous one $\mathcal{A}$:

\noindent\textbf{Constructing $\mathcal{C}$}. During training, we track the history of the last 5 predictions from $f$ for each snippet. Then, at the start of every epoch, we select $N$ snippets $\mathbf{x}_1,\ldots,\mathbf{x}_N$ with the least prediction variance, and the confident set $\mathcal{C}$ is given by $\{f(\mathbf{x}_i),y_i\}_{i=1}^N$. The rationale is that for the apparent normal or abnormal snippets (\eg, enclosed in red in Figure~\ref{fig:2}), their predictions tend to quickly converge to confident normal or abnormal with small predictive variance over time. This approach is empirically validated in Appendix, and we point out similar method shows promising results in~\cite{zhong2019graph}.

\noindent\textbf{Constructing $\mathcal{A}$}. The rest of the $M$ snippets have large prediction fluctuations, showing that $f$ is still uncertain about them.
They are collected as the ambiguous set $\mathcal{A} = \{\mathbf{x}_i\}_{i=1}^M$. Note that $\mathcal{A}$ is a set of features at this point, awaiting the next clustering step.

\subsection{Step 2: Clustering Ambiguous Snippets}

While the prediction from $f$ is ambiguous on $\mathcal{A}$, the feature distribution can still reflect the intrinsic differences between normal and abnormal snippets. Hence we aim to cluster $\mathcal{A}$ into 2 groups to distinguish them.
Specifically, we learn a cluster head $g$ that takes the snippet feature $\mathbf{x}\in \mathcal{A}$ as input and outputs the softmax-normalized probabilities for being in each of the 2 clusters. The head $g$ is trained in a pair-wise manner such that a pair of similar features have similar predictions from $g$ (\ie, from the same cluster), and vice versa for dissimilar. To accomplish this, we denote the pair-wise form of $\mathcal{A}$ based on cluster prediction from $g$ as:
\begin{equation}
    \mathcal{A}_g = \{ g(\mathbf{x}_i)^\intercal g(\mathbf{x}_j), \mathbbm{1}(\mathbf{x}_i \sim \mathbf{x}_j) \mid \mathbf{x}_i,\mathbf{x}_j \in \mathcal{A} \},
    \label{eq:2}
\end{equation}
where the dot-product is used to measure the prediction similarity, and $\mathbbm{1}(\cdot)$ is an indicator function that returns 1 if the cosine similarity between $\mathbf{x}_i,\mathbf{x}_j$ is larger than a threshold $\tau$ (\ie, $\mathbf{x}_i \sim \mathbf{x}_j$), and returns 0 otherwise. This allows us to train $g$ by minimizing $\textrm{BCE}(\mathcal{A}_g)$.

With the optimized $g$, each feature $\mathbf{x}_i$ in $\mathcal{A}$ is assigned a cluster label $y_i = \mathrm{argmax} \, g(\mathbf{x}_i)$ as the cluster with the highest predicted probability.
Next, we supervise $f$ by $\mathcal{A}$ to separate the clusters and form our overall objective.

\subsection{Step 3: Overall Objective}

Note that unlike the sample-wise supervision provided by labels in $\mathcal{C}$, \ie, whether a feature is normal or abnormal, the cluster labels in $\mathcal{A}$ only provide pair-wise supervision, \ie, whether a feature pair is from the same cluster.
Hence we supervise $f$ with $\mathcal{A}$ using a pair-wise loss: $f$ is trained to produce similar anomaly prediction on feature pairs with the same cluster label, and push away predictions for those in different clusters. This corresponds to minimizing $\textrm{BCE}(\mathcal{A}_f)$ with $\mathcal{A}_f$ based on the pair-wise prediction similarity of $f$:
\begin{equation}
    \mathcal{A}_f = \{ f(\mathbf{x}_i)^\intercal f(\mathbf{x}_j), \mathbbm{1}(y_i=y_j) \mid \mathbf{x}_i,\mathbf{x}_j \in \mathcal{A} \},
    \label{eq:3}
\end{equation}
where $f(\mathbf{x}_i)^\intercal f(\mathbf{x}_j)$ denotes the dot-product similarity of the binary probabilities (\ie, normal or abnormal)\footnote{While $f$ only outputs the probability of being abnormal as $p$, the probability of being normal is easily computed as $1-p$.} with slight abuse of notation. The overall objective of UMIL is given by:
\begin{equation}
    \mathop{\mathrm{min}}_{\theta,f,g} \overbrace{\textrm{BCE}(\mathcal{C})}^{\text{$\mathcal{C}$ supervision}} + \overbrace{\alpha \textrm{BCE}(\mathcal{A}_f)}^{\text{$\mathcal{A}$ supervision}} + \overbrace{\beta \textrm{BCE}(\mathcal{A}_g)}^{\text{Clustering in $\mathcal{A}$}},
    \label{eq:4}
\end{equation}
where $\alpha,\beta$ are trade-off parameters with ablations in Section~\ref{sec:Abla}. 
Hence in addition to the supervision from $\mathcal{C}$ as in MIL, $f$ in UMIL is additionally supervised by $\mathcal{A}$ to separate its 2 clusters identified by $g$ to remove the context bias in $\mathcal{C}$ (Figure~\ref{fig:2}). 
This unbiased objective allows us to train not only $f$, but also to fine-tune the backbone $\theta$ to get a tailored representation for VAD.

\noindent\textbf{Training and Testing}. Before training, the backbone $\theta$ is first pre-trained with MIL, and $f,g$ are randomly initialized. Then the models are trained with our proposed UMIL by iterating Algorithm~\ref{alg:1} until convergence. In testing, anomalies are labeled on the frame level. The model is evaluated with a non-overlapping sliding window of frames (\ie, each window of frames is a snippet) to predict anomaly whenever the window intersects with any anomaly frame.

\renewcommand{\algorithmicforall}{\textbf{for each}}
\begin{algorithm}[!t]
    \caption{UMIL Training (1 epoch)}
    \label{alg:1}
    \begin{algorithmic}[1]
        \STATE {\bfseries Input:} $N+M$ video snippets, backbone parameterized by $\theta$, classifier $f$ and cluster head $g$, batch size $b$.
        \STATE {\bfseries Output:} $\theta,f,g$ trained for 1 epoch.
        
        \STATE Compute $\{\mathbf{x}_i\}_{i=1}^{N+M}$ (features extracted by $\theta$)
        \STATE Update prediction history $\mathcal{H}\leftarrow \mathcal{H} \cup \{f(\mathbf{x}_i)\}_{i=1}^{N+M}$
        \STATE Construct $\mathcal{C}=\{f(\mathbf{x}_i),y_i\}_{i=1}^N, \mathcal{A}=\{\mathbf{x}_i\}_{i=1}^M$ from $\mathcal{H}$
        \REPEAT
            \STATE Sample a batch $\{f(\mathbf{x}_i),y_i\}_{i=1}^b$ from $\mathcal{C}$
            \STATE Compute $\textrm{BCE}(\mathcal{C})$ for the batch with Eq.~\eqref{eq:1}
            \STATE Sample a batch $\{\mathbf{x}_i\}_{i=1}^b$ from $\mathcal{A}$
            \STATE Assign $y_i\leftarrow\mathrm{argmax}\,g(\mathbf{x}_i)$ for $i\in\{1,\ldots,b\}$
            \STATE Construct $\mathcal{A}_g,\mathcal{A}_f$ with Eq.~\eqref{eq:2},~\eqref{eq:3} for the batch
            \STATE Compute $\textrm{BCE}(\mathcal{A}_g),\textrm{BCE}(\mathcal{A}_f)$
            \STATE Optimize $\theta,f,g$ with Eq.~\eqref{eq:4}
        \UNTIL{end of epoch}
    \end{algorithmic}
\end{algorithm}

\section{Experiments}
\label{sec:Exp}
\subsection{Datasets and Evaluation Metrics}
We conducted extensive experiments and ablations on two standard WSVAD evaluation datasets~\cite{sultani2018real,lv2021localizing}. As per standard in WSVAD, the training videos only have video-level labels, and the test videos have frame-level labels. Other details are given below:

\noindent\textbf{UCF-Crime}~\cite{sultani2018real} is a large-scale dataset that contains 1,900 untrimmed real-world outdoor and indoor surveillance videos. The total length of the videos is 128 hours, which contains 13 classes of anomalous events.
We follow the standard split: the training set contains 1,610 videos, and the test set contains 290 videos.

\noindent\textbf{TAD} dataset~\cite{lv2021localizing} contains real-world videos of traffic scenes with a total length of 25 hours and 1,075 average frames per video.
The videos contain more than 7 categories of anomalies that are common on roads.
The dataset is partitioned as a training set with 400 videos, and a test set with 100 videos.

\noindent\textbf{Evaluation Metrics}. Following previous works~\cite{sultani2018real,zhong2019graph}, we used the Area Under the Curve (AUC) of the frame-level ROC (Receiver Operating Characteristic) as the main evaluation metric for TAD and UCF-Crime. Intuitively, a larger AUC means a larger margin between the normal and abnormal snippet predictions, hence indicating a better anomaly classifier.
Inspired by Lv~\etal~\cite{lv2021localizing}, besides evaluating AUC on the overall test set with normal and abnormal videos, denoted as $\mathrm{AUC}_O$, we also computed the AUC on abnormal ones alone, denoted as $\mathrm{AUC}_A$.
The rationale is to remove normal videos where all snippets are normal (label 0), and keep only the abnormal ones with both kinds of snippets (label 0,1), which truly challenges a classifier's capability of localizing anomalies.
 
\subsection{Implementation Details}
\label{sec:ID}

\begin{figure}[t]
	\centering
	\includegraphics[width=\linewidth]{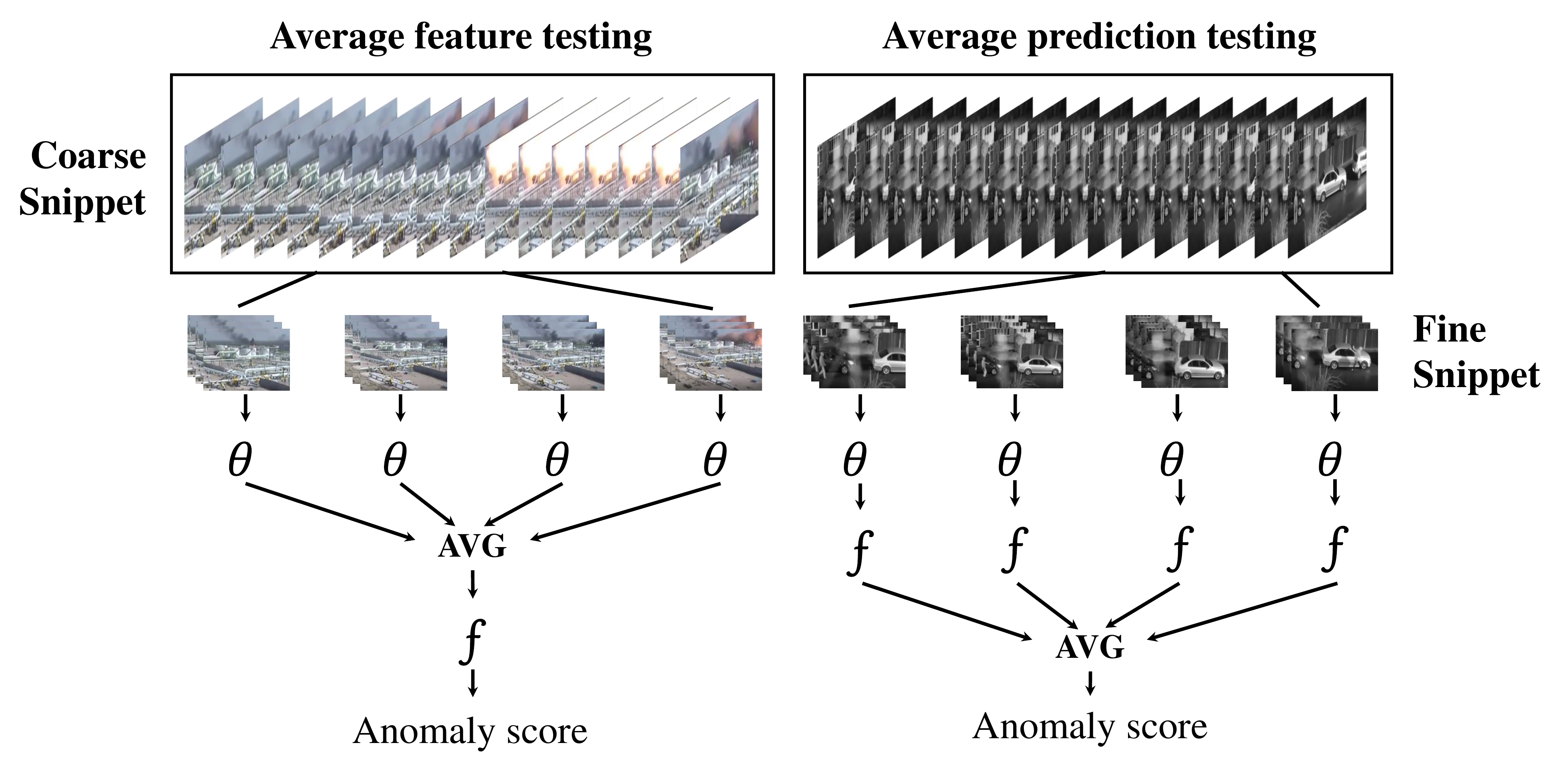}
    \vspace{-8mm}
	\caption{Average feature versus average prediction testing. $\theta,f$: the feature backbone and anomaly classifier, respectively.}
	\label{fig:pip}
    \vspace{-6mm}
\end{figure}

\noindent\textbf{Video Sequence Partition}. Existing works partition each video into multiple coarse snippets, and use the \emph{average feature} in each one as the input to their classifiers (Figure~\ref{fig:pip} left). However, we find that the subtle anomaly feature is often diluted by averaging features over the coarse snippets (see Appendix).
This has less impact on the traditional MIL compared to our UMIL, as MIL only leverages the confident snippets with apparent anomalies.
Therefore, in UMIL training, we used fine-grained snippets with one-second lengths. In testing, to generate the prediction for a coarse snippet, we used the \emph{average predictions} over the fine snippets inside the coarse one (Figure~\ref{fig:pip} right).

\noindent\textbf{Baseline}. We built a baseline to validate that the improvements of UMIL are indeed from the unbiased training scheme (Section~\ref{sec:Abla}), rather than the above testing scheme based on average predictions. Specifically, the baseline has exactly the same model design as UMIL, and we trained it with the MIL objective in Eq.~\eqref{eq:1} on fine snippets and tested it by averaging predictions. Hence the only difference between the baseline and UMIL is the training objective.

\noindent\textbf{Model Training}. We implemented the backbone $\theta$ with the X-CLIP-B/32 model~\cite{xclip} fine-tuned on Kinectics-400~\cite{carreira2017quo} to improve its capabilities in action recognition. We used the fully connected layer to implement the anomaly classifier $f$ and the cluster head $g$.
We trained our model with the AdamW optimizer~\cite{loshchilov2019decoupled} using an initial learning rate of $8$e-$6$, weight decay of $0.001$, and batch size of $8$.
We utilized the cosine annealing scheduler and warmed up the learning rate for 5 epochs.
Our UMIL model was pre-trained with MIL for 30 epochs, followed by 10 epochs of UMIL training.
We conducted all experiments on $4$ TITAN RTX GPUs.
We implement the max value scores as well as max margin scores~\cite{lv2021localizing} in $\mathcal{C}$ supervision of Eq~\ref{eq:4}.
We also incorporated entropy minimization as a standard auxiliary objective~\cite{liu2021cycle,long2018conditional}, and added the self-training loss, which leverages the learned unbiased anomaly classifier $f$ to generate accurate pseudo-labels on samples in the ambiguous set $\mathcal{A}$ for additional supervision. Details in Appendix.

\subsection{Main Results}
\label{sec:4.3}
\noindent\textbf{UCF-Crime and TAD}.
In Table~\ref{tab:ucf-crime}, we compared our UMIL with other state-of-the-art (SOTA) methods in both Unsupervised VAD (UVAD) and WSVAD. On UCF-Crime~\cite{sultani2018real}, UMIL achieves the best $\mathrm{AUC}_O$ and $\mathrm{AUC}_A$ among all the methods, with an improvement of +1.37\% and +1.3\%, respectively. UMIL also significantly outperforms all methods in TAD~\cite{lv2021localizing} by +3.3\% on $\mathrm{AUC}_O$ and +4.2\% on $\mathrm{AUC}_A$.

\begin{table}[t]
    \centering
    \scalebox{0.8}{
    \begin{tabular}{@{}c|c|c|c}
      \toprule\hline
        Category & Method         & $\mathrm{AUC}_O$ (\%) & $\mathrm{AUC}_A$ (\%)\\ 
      \hline\hline
      \multirow{6}{*}{\rotatebox{90}{UVAD}}
      & SVM Baseline   & 50.00  & 50.00     \\
      & Conv-AE~\cite{hasan2016learning}   & 50.60    & -   \\
      & Sohrab et al.~\cite{sohrab2018subspace}  & 58.50  & -  \\
      & Lu et al.~\cite{lu2013abnormal}  & 65.51  & -  \\
      & BODS~\cite{wang2019gods}           & 68.26  & -  \\
      & GODS~\cite{wang2019gods}           & 70.46  & -  \\ \hline
      \multirow{9}{*}{\rotatebox{90}{WSVAD}}
      & Sultani et al.~\cite{sultani2018real} & 75.41 &54.25    \\
      & Zhang et al.~\cite{zhang2019temporal}            & 78.66  & -    \\
      & Motion-Aware~\cite{zhu2019motion} & 79.10   & 62.18    \\
      & GCN-Anomaly~\cite{zhong2019graph} & 82.12  & 59.02    \\
      & Wu et al.~\cite{Wu2020not} & 82.44  & -    \\
      & RTFM~\cite{tian2021weakly}          & 84.30 & -   \\ 
      & WSAL~\cite{lv2021localizing}          & 85.38  & 67.38\\ 
      & \cellcolor{mygray}Baseline & \cellcolor{mygray}80.67  & \cellcolor{mygray}60.57 \\ 
      & \cellcolor{mygray}\textbf{UMIL}  & \cellcolor{mygray}\textbf{86.75}  & \cellcolor{mygray}\textbf{68.68} \\ \hline\bottomrule
    \end{tabular}%
    }
    \vspace{-2mm}
    \caption{Frame-level AUC performance on UCF-Crime. Best results in bold. $\mathrm{AUC}_O$ and $\mathrm{AUC}_A$ denote that the AUC computed on the overall test set and only abnormal test videos, respectively. ``UVAD'' and ``WSVAD'' under category denote Unsupervised VAD and Weakly-Supervised VAD, respectively.} 
    \label{tab:ucf-crime}
    \vspace{-4mm}
\end{table}

\noindent\textbf{Overall Observations}.
1) Notice that our baseline performs similarly (\eg, $\mathrm{AUC}_O$ on TAD) or even worse (\eg, 60.57\% versus 67.38\% on UCF-Crime $\mathrm{AUC}_O$) compared to existing MIL-based methods. This validates that the improvements from UMIL are not from the test scheme of averaging predictions. 
2) In particular, our improvement in $\mathrm{AUC}_A$ indicates that the superior performance of UMIL on $\mathrm{AUC}_O$ is not merely from easy normal videos, but also from improved capabilities to identify anomalous snippets in abnormal videos.
3) Moreover, on both datasets, WSVAD significantly improves over UVAD on $\mathrm{AUC}_O$, which empirically validates that detecting open-set anomalies in UVAD is ill-posed (Section~\ref{sec:intro}). However, the improvements in $\mathrm{AUC}_A$ are much smaller (\eg, 54.25\% over 50.00\% on UCF-Crime). This shows that the existing WSVAD methods are still biased toward the apparent normal/abnormal, causing many false positives and negatives on ambiguous snippets from the abnormal videos.
4) Our UMIL significantly improves the $\mathrm{AUC}_A$ over MIL (\eg, +4.2\% on TAD), which demonstrates the effectiveness of using ambiguous snippets in UMIL to learn an unbiased invariant classifier.
5) Interestingly, TAD tends to have larger $\mathrm{AUC}_O$ but lower $\mathrm{AUC}_A$, \eg, from UCF-Crime to TAD, UMIL's $\mathrm{AUC}_O$ is 6.2\% higher, but $\mathrm{AUC}_A$ is 2.8\% lower. The improved overall performance suggests that TAD has stronger context bias in the confident set, \ie, more apparent normal/abnormal snippets, and the dropped $\mathrm{AUC}_A$ indicates that it contains more subtle anomalies in the ambiguous snippets that are hard to detect and localize.
This also explains why our UMIL improves $\mathrm{AUC}_A$ more on TAD than UCF-Crime by incorporating ambiguous snippets to remove the context bias from the confident set.

\begin{table}[t]
    \centering
    \scalebox{0.95}{
    \begin{tabular}{@{}c|c|c|c}
      \toprule\hline
      Category       & Method         & $\mathrm{AUC}_O$ (\%) & $\mathrm{AUC}_A$ (\%)\\ \hline\hline
      \multirow{3}{*}{\rotatebox{90}{UVAD}}
      & SVM Baseline   & 50.00  & 50.00     \\
      & Luo~\etal~\cite{luo2017revisit}  & 57.89  & 55.84  \\
      & Liu~\etal~\cite{liu2018future}           & 69.13 & 55.38   \\ \hline
      \multirow{6}{*}{\rotatebox{90}{WSVAD}}
      & Sultani~\etal~\cite{sultani2018real} & 81.42 &55.97    \\
      & Motion-Aware~\cite{zhu2019motion} & 83.08  & 56.89    \\
      & GIG~\cite{lv2020global}          & 85.64 & 58.65   \\ 
      & WSAL~\cite{lv2021localizing}          & 89.64  & 61.66\\ 
      & \cellcolor{mygray}Baseline & \cellcolor{mygray}89.10 & \cellcolor{mygray}56.47 \\ 
      & \cellcolor{mygray}Ours & \cellcolor{mygray}{\textbf{92.93}} & \cellcolor{mygray}{\textbf{65.82}} \\ \hline\bottomrule
    \end{tabular}%
    }
    \vspace{-3mm}
    \caption{Frame-level AUC performance on TAD benchmark.} 
    \vspace{-3mm}
    \label{tab:tad}
\end{table}

\begin{table}[t]
\centering
\scalebox{0.75}{
\begin{tabular}{cccc|cc}
\toprule\hline
Baseline & ST & RTFM* & UMIL &  $\mathrm{AUC}_O$ (\%) - UCF  &  $\mathrm{AUC}_O$ (\%) - TAD\\ \hline \hline
\checkmark & & & & 80.67 & 89.10 \\
\checkmark & \checkmark & & & 82.01 & 90.80\\ \hline
\checkmark & \checkmark & \checkmark & & 83.45 & 91.28 \\ 
\checkmark & & & \checkmark & 83.66 & 91.74 \\ 
\cellcolor{mygray}\checkmark & \cellcolor{mygray}\checkmark & \cellcolor{mygray} & \cellcolor{mygray}\checkmark & \cellcolor{mygray}\textbf{86.75} & \cellcolor{mygray}\textbf{92.93}  \\ \hline
\bottomrule
\end{tabular}%
}
\vspace{-3mm}
\caption{Ablation studies of the components in UMIL on UCF-Crime and TAD. *: we re-implemented RTFM with our backbone and average-prediction-based testing scheme for fair comparison.}
\vspace{-3mm}
\label{tab:ablation}
\end{table}

\begin{table}[t]
\centering
\scalebox{1.0}{
\begin{tabular}{cccccc}
\toprule\hline
Threshold(\%) & 10 & \cellcolor{mygray}\textbf{30} & 50 & 70 & 90  \\ \hline \hline
$\mathrm{AUC}_O$ (\%) - UCF & 86.8 & \cellcolor{mygray}\textbf{86.8} & 85.9 & 84.3 & 83.1 \\ 
$\mathrm{AUC}_O$ (\%) - TAD & 92.7 & \cellcolor{mygray}\textbf{93.0} & 92.8 & 91.5 & 91.1 \\ \hline
\bottomrule
\end{tabular}%
}
\vspace{-3mm}
\caption{Ablation on the threshold to divide the confident/ambiguous snippet set on UCF-Crime and TAD.}
\label{tab:thre}
\vspace{-5mm}
\end{table}
\subsection{Ablations}
\label{sec:Abla}

\noindent\textbf{Components}. Our approach has 2 main components: 1) the self-training objective; 2) the UMIL objective in Eq.~\eqref{eq:4}. We validate the effectiveness of each component in Table~\ref{tab:ablation} with $\mathrm{AUC}_O$. All ablations in the table are on the equal ground---using average prediction instead of average feature for anomaly detection (\ie, Baseline). By comparing the first two lines, we observe that self-training can improve $\mathrm{AUC}_O$ from $80.67\%$ to $82.01\%$ on UCF-crime and $89.10\%$ to $90.80\%$ on TAD. To independently evaluate the effectiveness of UMIL objective, we re-implement the SOTA RTFM~\cite{tian2021weakly} using our backbone and add the self-training objective, namely RTFM*. The result is listed in line 3. Our UMIL in line 4 still significantly outperforms RTFM* (+$3.3\%$ on UCF-crime and +$1.7\%$ on TAD), hence validating the effectiveness of our unbiased learning objectives.
\begin{figure}[t]
	\centering
	\includegraphics[width=0.4\textwidth]{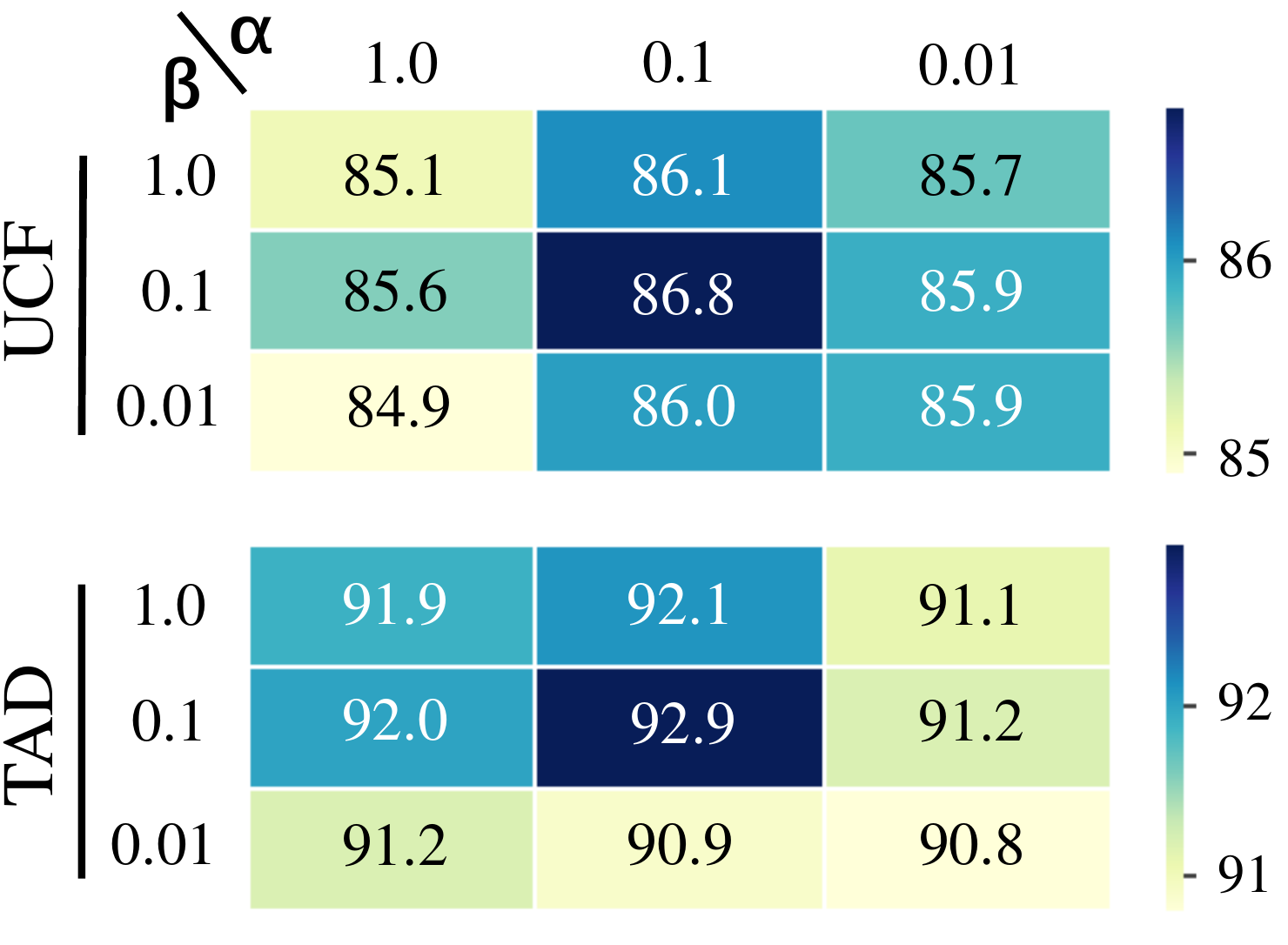}
    \vspace{-4mm}
	\caption{Ablations on the trade-off parameters.}
	\label{fig:cm}
    \vspace{-4mm}
\end{figure}

\noindent\textbf{Confident Threshold}. We then conducted experiments to analyze the effects of the variance threshold for dividing confident and ambiguous snippets as in Section~\ref{sec:step1}.
Specifically, we selected $k$ (\%) training snippets with the minimum variance on their prediction history with varying $k$ as in Table~\ref{tab:thre}. Overall the threshold is easy to determine, \ie, 10-50\% is a reasonable range with 30\% being the best.

\noindent\textbf{Trade-off Parameters}. Recall that we use $\alpha$ and $\beta$ in Eq.~\eqref{eq:4} as the trade-off for the supervision from the ambiguous set $\mathcal{A}$ and clustering, respectively. We empirically find in Figure~\ref{fig:cm} that $\alpha,\beta=0.1$ are suitable across the two datasets, hence we used this setting in the experiments by default. In general, the choice of $\alpha$ depends on the strength of the context bias in the confident set, \eg, TAD has strong bias as analyzed in Section~\ref{sec:4.3}, which cannot be overcome with a small $\alpha$ (\eg, $\alpha$=0.01 has low performance).

\begin{figure}
    \centering
    \footnotesize
    \scalebox{1.05}{
    \begin{subfigure}[t]{0.23\textwidth}
         \includegraphics[width=\textwidth]{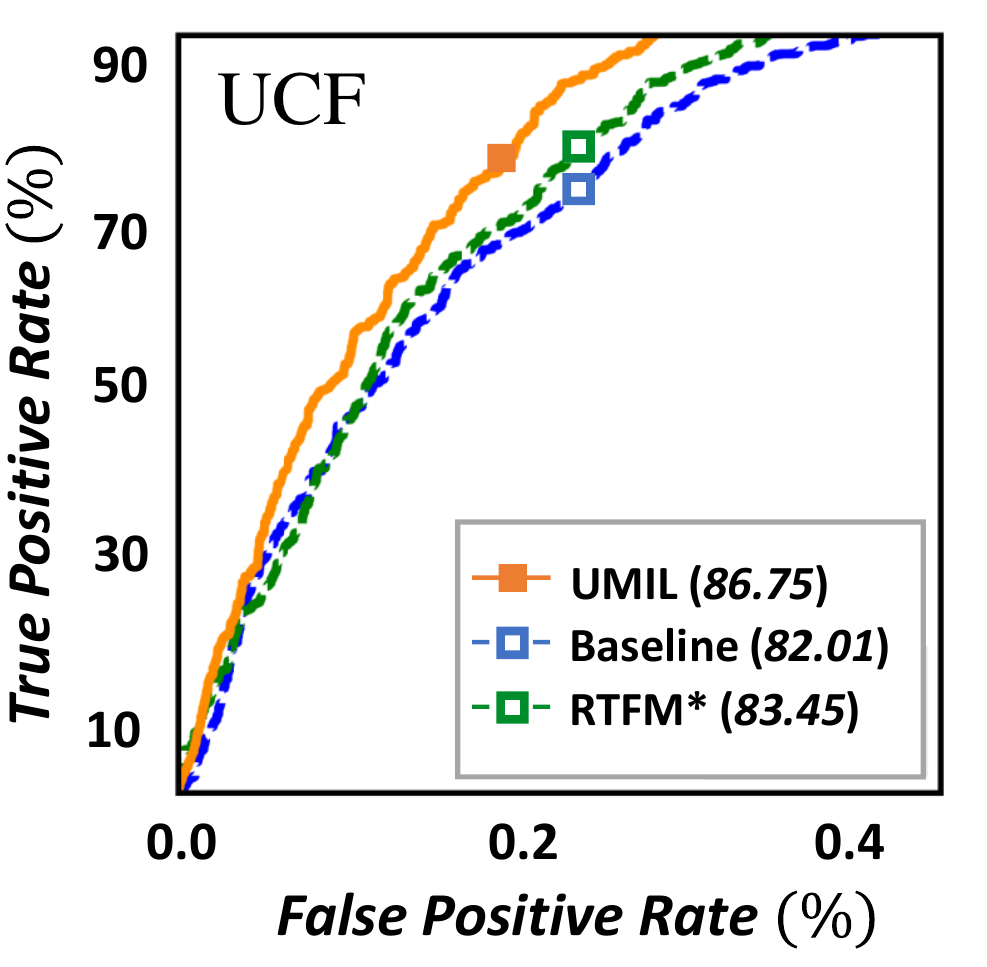}
         \phantomcaption
         \label{fig:roca}
    \end{subfigure}
    \begin{subfigure}[t]{0.23\textwidth} 
         \includegraphics[width=\textwidth]{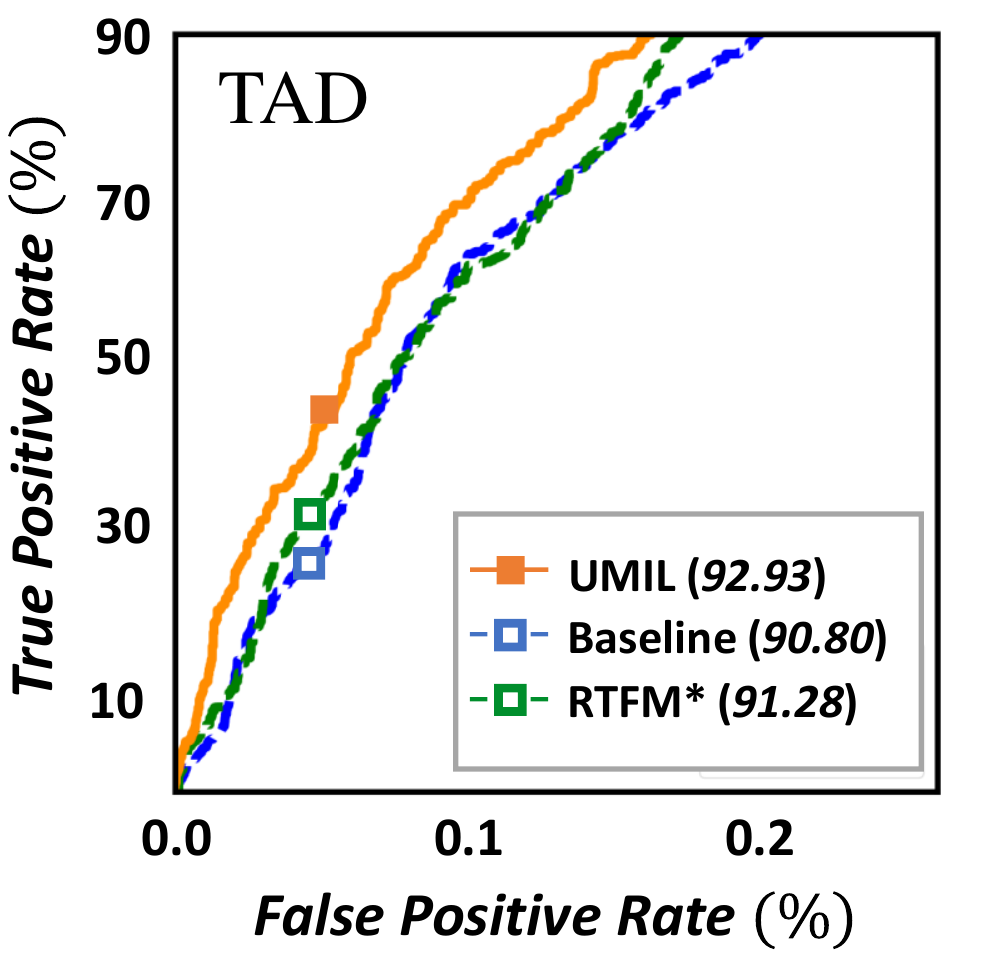}
         \phantomcaption
         \label{fig:rocb}   
    \end{subfigure}}
    \vspace{-8mm}
    \caption{ROC curves on UCF and TAD. Note that we only show part of the curves for visual clarity, as the other part of the methods have a large overlap when the true positive rate approaches 100\%.}
    \label{fig:roc}
    \vspace{-6mm}
\end{figure}

\begin{figure}[t]
	\centering
	\includegraphics[width=\linewidth]{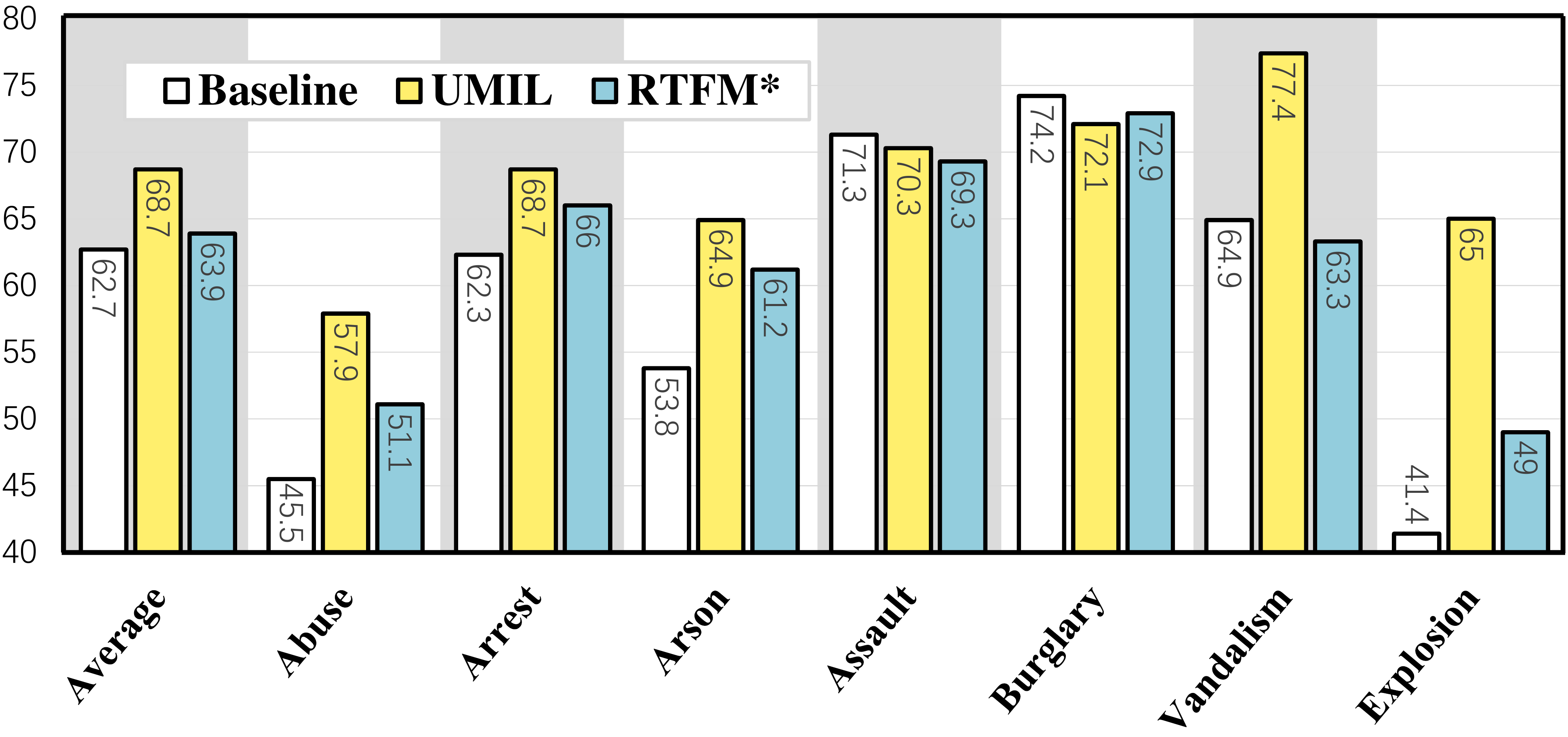}
    \vspace{-6mm}
	\caption{Class-wise $\mathrm{AUC}_A$ of three methods on UCF-Crime.}
	\label{fig:hist}
    \vspace{-7mm}
\end{figure}

\noindent\textbf{Class-wise AUC}. On UCF-Crime dataset, the class of anomaly in each test video is given. This allows us to plot the class-wise $\mathrm{AUC}_A$ to examine models' capabilities to detect subtle abnormal events. In Figure~\ref{fig:hist}, we compared UMIL with baseline and RTFM*, where ``Average'' shows the overall $\mathrm{AUC}_A$ and the rest shows the class-wise one.
We have the following observation:
1) Both of the two MIL-based methods perform well on human-centric anomaly classes with drastic motions, \eg, ``Assault'' and ``Burglary''. These classes correspond to apparent anomalies as the backbone expresses the human action feature well (fine-tuned on the action recognition Kinetics400 dataset\cite{carreira2017quo}).
2) However, we notice that they easily fail to distinguish anomalies with subtle motions, \eg, ``Arson'' and ``Vandalism'', as well as non-human-centric anomalies, \eg, ``Explosion''. These classes correspond to ambiguous anomalies discarded by the biased training in MIL.
3) Our UMIL performs similarly on the above apparent anomaly classes and much better on the other subtle anomalies, which largely contributes to the superior anomaly detection and localization performance.
Overall, observation 1 and 2 empirically verifies the biased prediction situation of MIL in Figure~\ref{fig:abstract} and Figure~\ref{fig:2}. In contrast, our UMIL convincingly improves the performance on ambiguous anomalies with almost no sacrifice on the confident ones, which validates the effectiveness of our approach, \ie, identifying the invariance between the two types of anomalies to remove the bias in MIL.

\begin{figure*}[t!]
	\centering
    \vspace{-4mm}
	\includegraphics[width=1\textwidth]{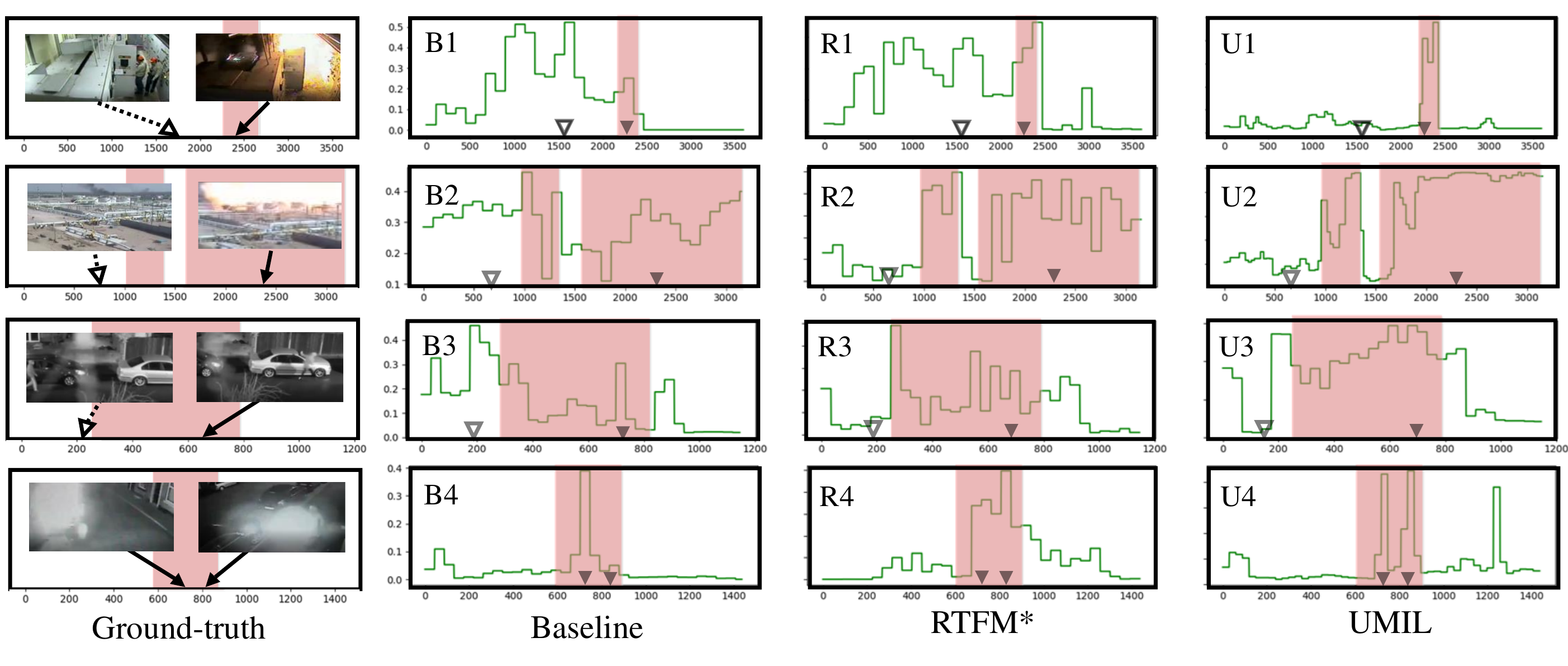}
    \vspace{-10mm}
	\caption{Visualization cases of ground-truth and anomaly score curves of various approaches. The white and black triangles denote the location of the normal and abnormal frame displayed on the left, respectively. The green curves represent the anomaly predictions of various methods. The pink background corresponds to the ground-truth abnormal regions.}
	\label{fig:cases}
    \vspace{-4mm}
\end{figure*}

\noindent\textbf{ROC Curve}. In Figure~\ref{fig:roc}, we draw the ROC Curve on the overall test set for our baseline, the re-implemented RTFM* and UMIL, which shows the true and false positive rate for detecting anomaly on a sweeping threshold over the predictions. VAD is evaluated using the area under this curve to demonstrate the overall separation of normal and abnormal snippet predictions. However, when applying a detector for real-world usage, we need to choose a specific threshold (\eg, with a maximum tolerable false positive rate). We observe from Figure~\ref{fig:roc} that our UMIL outperforms the two MIL baselines in every inch, which further shows the strength of our proposed unbiased training.

\noindent\textbf{Qualitative Analysis}. In Figure~\ref{fig:cases}, we show the continuous predictions of anomaly probabilities from our baseline, RTFM*, and our UMIL on 4 test videos on UCF-crime. We summarize the observations:
1) For the MIL baseline (2nd column), we observe that it assigns a larger probability on the pre-explosion snippets from B1 and B2 (top two videos), \eg, workers performing maintenance and snippets with smoke, yet the actual explosion may have a lower prediction (\eg, comparing the height of the green lines on the white and black triangle locations). Similarly, on B3, the running person (white triangle) triggers a larger anomaly prediction than the actual vandalism (black triangle). This further illustrates the biased prediction problem in MIL.
2) RTFM (3rd column) uses feature magnitudes to assist anomaly detection by assuming anomalous snippets have larger magnitudes, which indeed improves over the baseline sometimes, \eg, R2 is no longer biased to smoke. However, its assumption has no guarantee to hold and hence the failure on subtle anomalies persists, \eg, false alarm in R1 white triangle location and low prediction in R3 black triangle location.
3) In contrast, our UMIL localizes the anomalies accurately in U1-U3, \eg, having consistently high scores in the pink areas, which holds its ground on the name ``unbiased''.
4) In the 4th video, however, RTFM's prediction in the pink area is more consistent than ours. By inspecting the frames on the left, we realize that the two peaks in the pink area of U4 correspond to the burning fire and the running suspect caught on fire. Hence UMIL's prediction is reasonable and sufficient for triggering the alarm on the first peak.

\noindent\textbf{Computational Efficiency}. Lastly, we investigated the speed of the proposed model.
For inference, our method processes a 5-frame clip in $0.003$ seconds on a Nvidia 2080Ti GPU.
Notably, this is almost $80 \times$ faster than the SOTA RTFM~\cite{tian2021weakly}, which spends 0.76 seconds to process a 16-frame clip on Nvidia 2080Ti.
Thanks to our unbiased training scheme, we can fine-tune the backbone to learn a WSVAD-tailored representation, which achieves even better performance than existing SOTA.
This also shows the promising future of UMIL in real-time applications.
\section{Conclusion}
\label{sec:Con}
In this work, we presented an Unbiased Multiple Instance Learning (UMIL) scheme that learns an unbiased anomaly classifier and a tailored representation for Weakly Supervised Video Anomaly Detection (WSVAD). Specifically, the existing MIL training scheme suffers from the context bias by only training on the confident set containing apparent normal/abnormal video snippets. We replace it with an unbiased one---seeking the invariant predictor that simultaneously distinguishes the normal/abnormal snippets in the confident set, and separates the two unsupervised clusters in the rest ambiguous snippets. Hence the context bias that fails among the ambiguous ones is removed. Our approach is empirically validated by the state-of-the-art performance and extensive ablations on standard WSVAD benchmarks. In future, we will seek additional prior beyond unsupervised clustering to discover the intrinsic differences between the ambiguous normal and abnormal snippets and adopt principled representation learning paradigm (\eg, disentanglement) to highlight the anomaly features.
\section{Acknowledge}
The author gratefully acknowledges the support of the A*STAR under its AME YIRG Grant (Project No.A20E6c0101), the Lee Kong Chian (LKC) Fellowship fund awarded by Singapore Management University, AI Singapore AISG2-RP-2021-022, the Postgraduate Research \& Practice Innovation Program of Jiangsu Province, the National Natural Science Foundation of China (Grants No.62072244), the Natural Science Foundation of Shandong Province (Grant No.ZR2020LZH008). This work was also supported in part by State Key Laboratory of High-end Server \& Storage Technology.
\section{Appendix}
This section is organized as follows:

\begin{itemize}[leftmargin=+0.1in,itemsep=2pt,topsep=0pt,parsep=0pt]
    \item Section~\ref{sec:LO} provides more details about our training objectives. We detail the implementation of the self-training used in our experiments: FixMatch~\cite{sohn2020fixmatch}.
    \item Section~\ref{sec:DIS} explains about how the feature clustering boosts the UMIL during training. 
    \item Section~\ref{sec:AE} shows more comparisons and standard deviations on UCF-crime~\cite{sultani2018real}, and TAD~\cite{lv2021localizing}. In particular, we first discuss the statics of anomaly events in UCF-crime in Section~\ref{sec:pp}, and then provide more experimental results of the proposed UMIL.
    \item Section~\ref{sec:roc} gives the full version of ROC curves on various benchmarks. This is a supplement to Figure.~6 in the manuscript.
    \item Codes are also provided, which include the training and testing scripts on the two classic datasets. The setup instructions and commands used in our experiments are included in the \texttt{README.md} file.
\end{itemize}

\section{Loss Objectives}
\label{sec:LO}
In this section, we give the details of the self-training objective $\mathcal{L}_{st}$ used in the MIL pre-training and UMIL training, as in Eq.(1) and Eq.(4) in the manuscript, then the overall MIL pre-training objective is derived as the following:
\begin{equation}
    \mathcal{L}_{mil} =  \textrm{BCE}(\mathcal{C}) + \lambda \mathcal{L}_{st} ,
    \label{eq:1}
\end{equation}
where $\lambda$ stands for the balance weight.
The overall objective of UMIL is derived correspondingly.
Note that self-training strategy is an important approach popular in domain adaption~\cite{french2017self,long2018conditional,liu2021cycle,zou2018unsupervised,kumar2020understanding}.
In this work, we introduce self-training to boost feature learning in WSVAD by incorporating data augmentation of \textbf{FixMatch}~\cite{sohn2020fixmatch}.
Specifically, along with the training of MIL, we generate pseudo labels with original video snippet data and seek to minimize the entropy between the predictions of augmented data as well as original data.
Given the pair of feature $\textbf{x}$ and $\textbf{x}'$ from the original data and random augmentation data, respectively, the \textbf{FixMatch}-driven self-training loss derives as:
\begin{equation}
    \mathcal{L}_{st} = \mathbbm{1} (\textrm{argmax}f(\textbf{x}) > \delta) \textrm{BCE} (\textbf{x}',\textrm{argmax}f(\textbf{x})),
    \label{eq:2}
\end{equation}
here, $\mathbbm{1}$ represents the indicator function that returns 1 if the condition is met, $\delta$ is the confident threshold, and $\textrm{BCE}$ corresponds to the binary cross entropy mentioned in the manuscript.
In the experiments, we used grid searching for finding a proper $\delta$. 
The results are listed in Section~\ref{sec:AE}.
\begin{figure}[t]
    \centering
    \footnotesize
    \begin{subfigure}[t]{0.23\textwidth}
         \includegraphics[width=\textwidth]{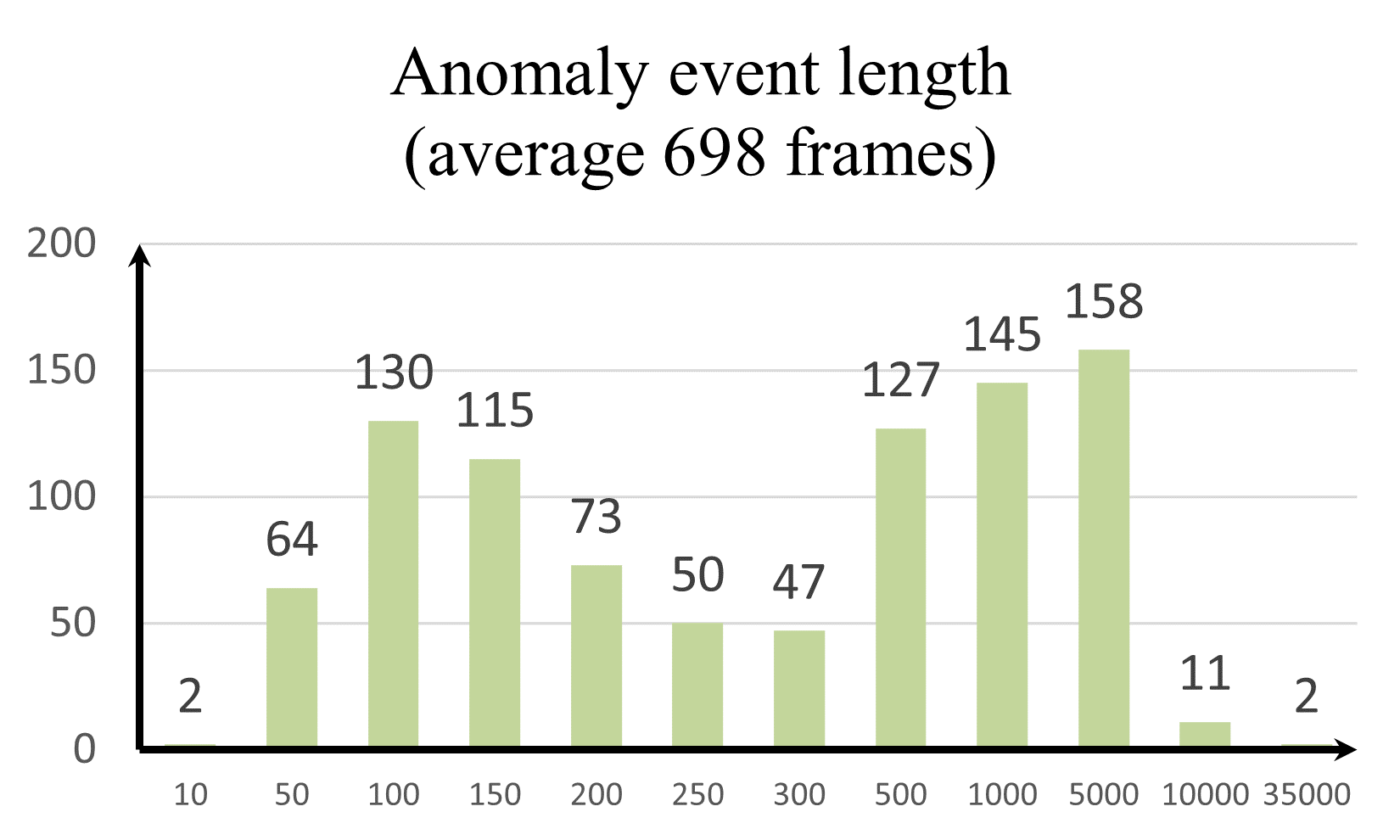}
         \caption{}
         \label{fig:1a}
    \end{subfigure}
    \begin{subfigure}[t]{0.23\textwidth} 
         \includegraphics[width=\textwidth]{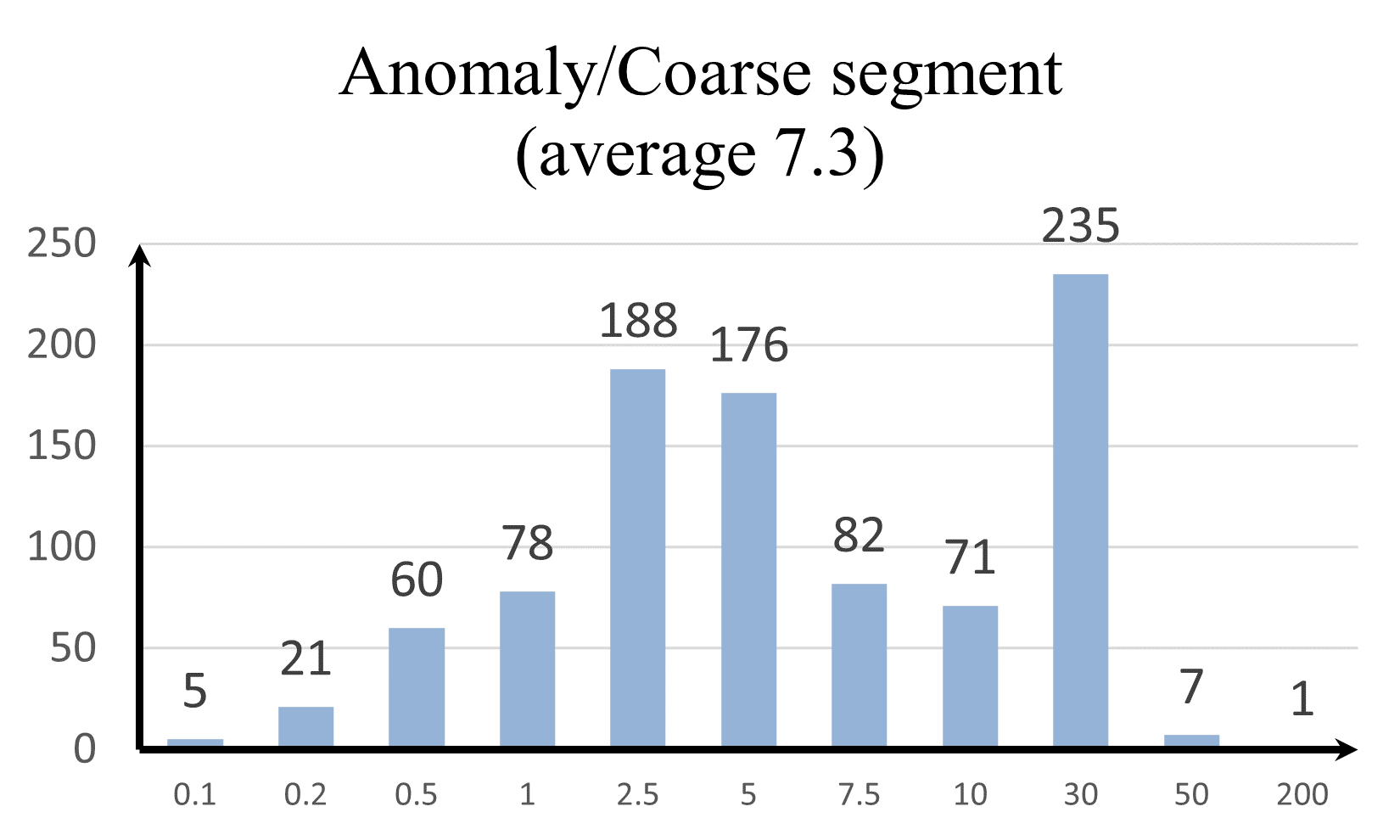}
         \caption{}
         \label{fig:1b}   
    \end{subfigure}
	\caption{(a) The statistics of anomaly event length and (b) the ratio of anomaly event length to coarse snippet length.}
	\label{fig:ano_rate}
\end{figure}

\section{Discussion on clustering}
\label{sec:DIS}
During the training of UMIL, wrong clustering can bring risks. However, the modern pre-trained backbones (\eg, CLIP) capture rich prior knowledge, such that the intrinsic difference between normal and abnormal snippets is sufficiently expressed in the feature space. This ensures that 1) during clustering, the normal/abnormal snippets can be separated into different clusters; and 2) combining $\mathcal{C}$- and $\mathcal{A}$- supervision in Eq. 4 leads to a classifier using true anomaly features instead of context for prediction (\eg, vertical black line in Figure 2). 
In future work, we will explore other prior knowledge or inductive bias to further separate normal/abnormal snippets.

\section{Additional Experiments}
\label{sec:AE}

\subsection{Pre-processing Analysis}
\label{sec:pp}
In this section, we first analyze the rationality in the pre-processing step of previous WSVAD approaches. 
As mentioned in Section 4.2 of the manuscript, existing works follow the average feature pipeline. 
They first divide video sequences into multiple coarse snippets, \eg 1 video 32 snippets, then take the \textit{snippet-level average features} as inputs into anomaly detectors. 
However, real-world anomalies are extremely rare and short in time. 
The subtle anomaly events are easily diluted or even covered by normal patterns after the spatio-temporal pooling operation.

To better analyze the problem, we annotate the large training set of UCF-crime~\cite{sultani2018real}. 
In detail, 5 trained annotators are involved in the process and the final labels are generated by averaging the results.
In Figure~\ref{fig:ano_rate}, we depict the statistics of the anomaly events' length. 
The average length of anomaly events is about 698 frames (extracted from videos with 30FPS), compared with an average coarse snippets' length of 200 frames. 
Note that  coarse snippets' length is obtained by dividing each video into 32 snippets, which is widely used as the default in existing works~\cite{sultani2018real,lv2021localizing,zhu2019motion}.
We also depict the ratio of anomaly event length to coarse snippet length in Figure~\ref{fig:1b}.
As is shown, there are 164 out of 925 anomaly events whose length ratios are less than 1. 
It means that these anomaly events are short than the coarse snippet.
More importantly, the length ratios of 86 anomaly events are less than 0.5.
Considering the anomalies only take place in a small part of whole frames, the anomaly information is inevitably concealed in the spatio-temporal feature pooling process, which is hurtful for video anomaly detection.

\subsection{Feature Fine-tuning Analysis}
When the backbone is loaded with pre-trained weights on kinetics 400 and frozen during UMIL training, the performance will drop from $86.75\%$ (with fine-tuning) to $83.44\%$ (frozen) on UCF, and $92.93\%$ to $90.71\%$ on TAD. This validates that fine-tuning in UMIL enables learning a representation tailored for WSVAD, which is beneficial for anomaly detection.

\subsection{Self-training In UMIL}
In this work, we use the learned anomaly classifier to generate pseudo-labels on samples in the ambiguous set $\mathcal A$.
Consequently, the ambiguous samples, which are largely neglected in existing MIL, can further participate in our UMIL with pseudo-labels.
When the self-training loss is removed, the results of UMIL are $83.66\%$ ($\downarrow3.09\%$) on UCF and $91.74\%$ ($\downarrow 1.19\%$) on TAD. 
This validates the effectiveness of the self-training loss. Further experimental analysis of the Self-training can be found in the following.

\subsection{Confident Threshold in Self-training}
\label{sec:ct}
In Table~\ref{tab:conf}, we list the results of adding self-training to MIL baseline model with varying confident threshold $\delta$. By increasing the confident threshold, less and highly confident samples are involved in the objective of self-training. As is shown in the table, $0.8$ is a suitable threshold that the self-training tool obtains good results on both datasets. When the threshold is up to 1, few samples will be selected leading to the ineffectiveness of self-training.
\begin{table}[t]
\centering
\scalebox{0.9}{
\begin{tabular}{ccccccc}
\toprule\hline
Threshold(\%)  & 0.3 & 0.5 & 0.7 & \cellcolor{mygray}\textbf{0.8} & 0.9 & 1.0 \\ \hline \hline
$\mathrm{AUC}_O$ (\%) - UCF & 80.9 & 81.2 & 81.9 & \cellcolor{mygray}\textbf{82.0} & 81.5 & 80.7\\ 
$\mathrm{AUC}_O$ (\%) - TAD & 89.0 & 90.1 & 90.5 & \cellcolor{mygray}\textbf{90.8} & 90.1 & 89.1 \\ \hline
\bottomrule
\end{tabular}%
}
\caption{Ablation on the Confident threshold in self-training based on MIL model on UCF-Crime and TAD.}
\label{tab:conf}
\end{table}

\subsection{Similarity Threshold in Clustering}
\label{sec:st}
The cluster component alone is for separating normal/abnormal snippets in the ambiguous set $\mathcal{A}$ as two clusters. 
It doesn't directly benefit the learning of anomaly classifier $f$. 
If the $\mathcal{A}$-supervision is removed, $f$ will be trained only on confident normal/abnormal snippets, and our approach will basically reduce to the existing MIL with similar performance.
We also conducted experiments to analyze the effect of varying similarity thresholds in clustering.
The experimental results are listed in Table~\ref{tab:simi}.
As we can see, the performance is insensitive to the change of the threshold in cosine similarity in Eq.~(2) of the manuscript.
Because the clustering property is acquired along with the feature fine-tuning of the backbone.
Then $0.8$ is chosen as the default similarity threshold in clustering.
\begin{table}[t]
\centering
\scalebox{1.0}{
\begin{tabular}{cccccc}
\toprule\hline
Threshold(\%) & 0.5 & 0.6 & 0.7 & \cellcolor{mygray}\textbf{0.8} & 0.9\\ \hline \hline
$\mathrm{AUC}_O$ (\%) - UCF & 86.4 & 86.6 & 86.8 & \cellcolor{mygray}\textbf{86.8} & 86.6 \\ 
$\mathrm{AUC}_O$ (\%) - TAD & 92.5 & 92.7 & 92.8 & \cellcolor{mygray}\textbf{93.0} & 92.9 \\ \hline
\bottomrule
\end{tabular}%
}
\caption{Ablation on the similarity threshold in clustering on UCF-Crime and TAD.}
\label{tab:simi}
\end{table}

\subsection{Confident Sample Selection Strategy}
\label{sec:css}
In this section, we also conducted experiments to compare \textit{Historical Variance} with \textit{Max Confidence} in the confident sample selection strategy.
Specifically, we select the top k (\%) abnormal and normal snippets with maximum confidence in abnormal videos as the confident set $\mathcal{C}$.
As we can see, the best AUC performances of \textit{Historical Variance} ($86.8\%$ for UCF-crime and $93.0\%$ for TAD) are superior to those of \textit{Max Confidence} ($85.9\%$ for UCF-crime and $92.2\%$ for TAD).
As mentioned in the manuscript (Section 3.2 Step 1), the predictions of the `easy' normal or abnormal snippets tend to quickly converge to confident normal or anomaly with small variance over time.
As a result, collecting historical information of score variance is a better choice for distinguishing confident and ambiguous samples.
\begin{table}[t]
\centering
\scalebox{1.0}{
\begin{tabular}{cccccc}
\toprule\hline
Threshold(\%) & 10 & 30 & 50 & 70 & 90  \\ \hline \hline
$\mathrm{AUC}_O$ (\%) - UCF & \cellcolor{mygray}\textbf{85.9} & 85.7 & 85.3 & 84.6 & 83.8 \\ 
$\mathrm{AUC}_O$ (\%) - TAD & 92.1 & \cellcolor{mygray}\textbf{92.2} & 92.0 & 91.4 & 90.8 \\ \hline
\bottomrule
\end{tabular}%
}
\caption{Ablation on the max confident threshold to divide the confident/ambiguous snippet set on UCF-Crime and TAD.}
\label{tab:max}
\end{table}

\subsection{Comparison on ShanghaiTech}
We also added experiment on ShanghaiTech benchmark \cite{liu2018future}. We failed to implement our method on Ped2, due to the absence of data splits. We achieved comparative performance with existing SOTAs as shown below, and believed that there is potential for further improvements given more time. Additionally, the high accuracy on this dataset indicates that it may contain mainly apparent anomalies, which explains why MIL-based methods already perform well (Section 3.1).
\begin{table}[h]
    \centering
    \small
    \scalebox{1.2}{
    \begin{tabular}{cccccc}
        \hline
        \textbf{Method} & GCN & RTFM & Baseline & Ours\\ 
        \hline
        \textbf{AUC(\%)} & 84.44 & 97.21 & 95.20 & 96.78\\ 
        \hline
    \end{tabular}}
\end{table}

\section{Visualization of ROC Curves}
\label{sec:roc}
\begin{figure}
    \centering
    \footnotesize
    \scalebox{1.05}{
    \begin{subfigure}[t]{0.23\textwidth}
         \includegraphics[width=\textwidth]{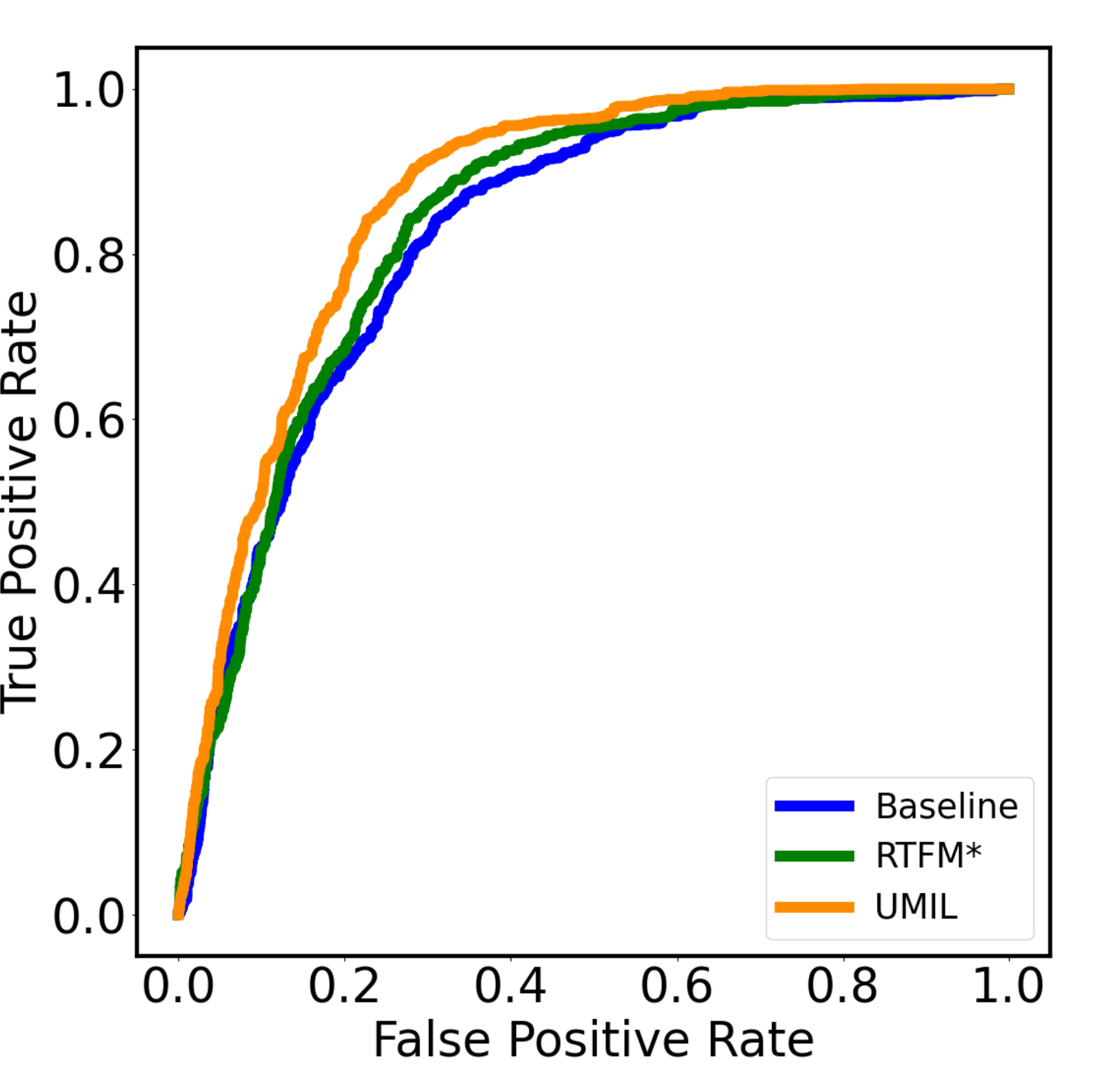}
         \phantomcaption
         \label{fig:roca}
    \end{subfigure}
    \begin{subfigure}[t]{0.23\textwidth} 
         \includegraphics[width=\textwidth]{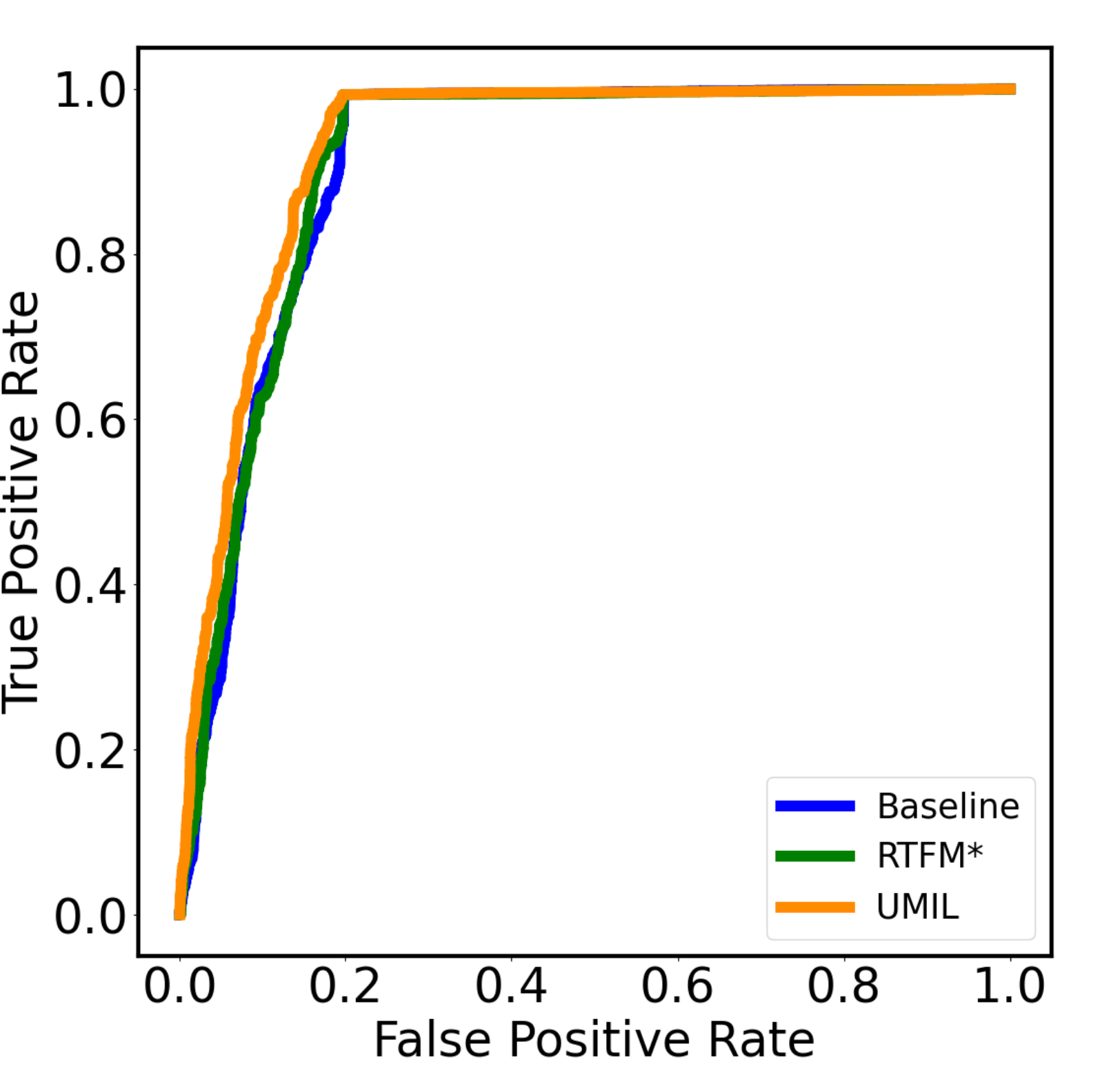}
         \phantomcaption
         \label{fig:rocb}   
    \end{subfigure}}
    \caption{Full version of the ROC curves on UCF (left) and TAD (Right).}
    \label{fig:roc}
\end{figure}

{\small
\bibliographystyle{ieee_fullname}
\bibliography{egbib}
}

\end{document}